\documentclass{article}

\usepackage{microtype}
\usepackage{graphicx}
\usepackage{subfigure}
\usepackage{booktabs} 
\usepackage{amsmath}
\usepackage{amssymb}
\usepackage{comment}
\usepackage{multirow}
\usepackage{threeparttable}
\usepackage{caption}
\usepackage{hyperref}

\usepackage{multicol}
\usepackage[accepted]{icml2024}

\usepackage{soul}

\definecolor{moh_colour}{RGB}{255, 204, 204}

\icmltitlerunning{Applying t-Distributions to Explore Accurate and Efficient Formats for LLMs}

\begin{document}

\twocolumn[
\icmltitle{\texorpdfstring{Learning from Students: Applying t-Distributions \\ to Explore Accurate and Efficient Formats for LLMs}{Learning from Students: Applying t-Distributions to Explore Accurate and Efficient Formats for LLMs}}


\begin{icmlauthorlist}
\icmlauthor{Jordan Dotzel}{cornell,google}
\icmlauthor{Yuzong Chen}{cornell}
\icmlauthor{Bahaa Kotb}{cornell}
\icmlauthor{Sushma Prasad}{google}
\icmlauthor{Gang Wu}{google}
\icmlauthor{Sheng Li}{google}
\icmlauthor{Mohamed S. Abdelfattah}{cornell}
\icmlauthor{Zhiru Zhang}{cornell}

\icmlcorrespondingauthor{Jordan Dotzel}{dotzel@cornell.edu}

\end{icmlauthorlist}


\icmlaffiliation{cornell}{School of ECE, Cornell University}

\icmlaffiliation{google}{Google}

\icmlkeywords{Machine Learning, ICML}

\vskip 0.3in
]


\printAffiliationsAndNotice{} 

\begin{abstract}
The increasing size of large language models (LLMs) traditionally requires low-precision integer formats to meet strict latency and power demands.
Yet recently, alternative formats such as Normal Float (NF4) have increased model accuracy at the cost of increased chip area.
In this work, we first conduct a large-scale analysis of LLM weights and activations across 30 networks and conclude that most distributions follow a Student's t-distribution.
We then derive a new theoretically optimal format, Student Float (SF4), that improves over NF4 across modern LLMs, for example increasing the average accuracy on LLaMA2-7B by 0.76\% across tasks.
Using this format as a high-accuracy reference, we then propose augmenting E2M1 with two variants of \textit{supernormal} support for higher model accuracy.
Finally, we explore the quality and efficiency frontier across 11 datatypes by evaluating their model accuracy and hardware complexity.
We discover a Pareto curve composed of INT4, E2M1, and E2M1 with supernormal support, which offers a continuous tradeoff between model accuracy and chip area.
For example, E2M1 with supernormal support increases the accuracy of Phi-2 by up to 2.19\% with 1.22\% area overhead, enabling more LLM-based applications to be run at four bits.
The supporting code is hosted at \href{https://github.com/cornell-zhang/llm-datatypes}{https://github.com/cornell-zhang/llm-datatypes}.
\end{abstract}

\section{Introduction}
\label{sec:intro}

\begin{figure}[t]
  \centering
   \includegraphics[width=\linewidth]{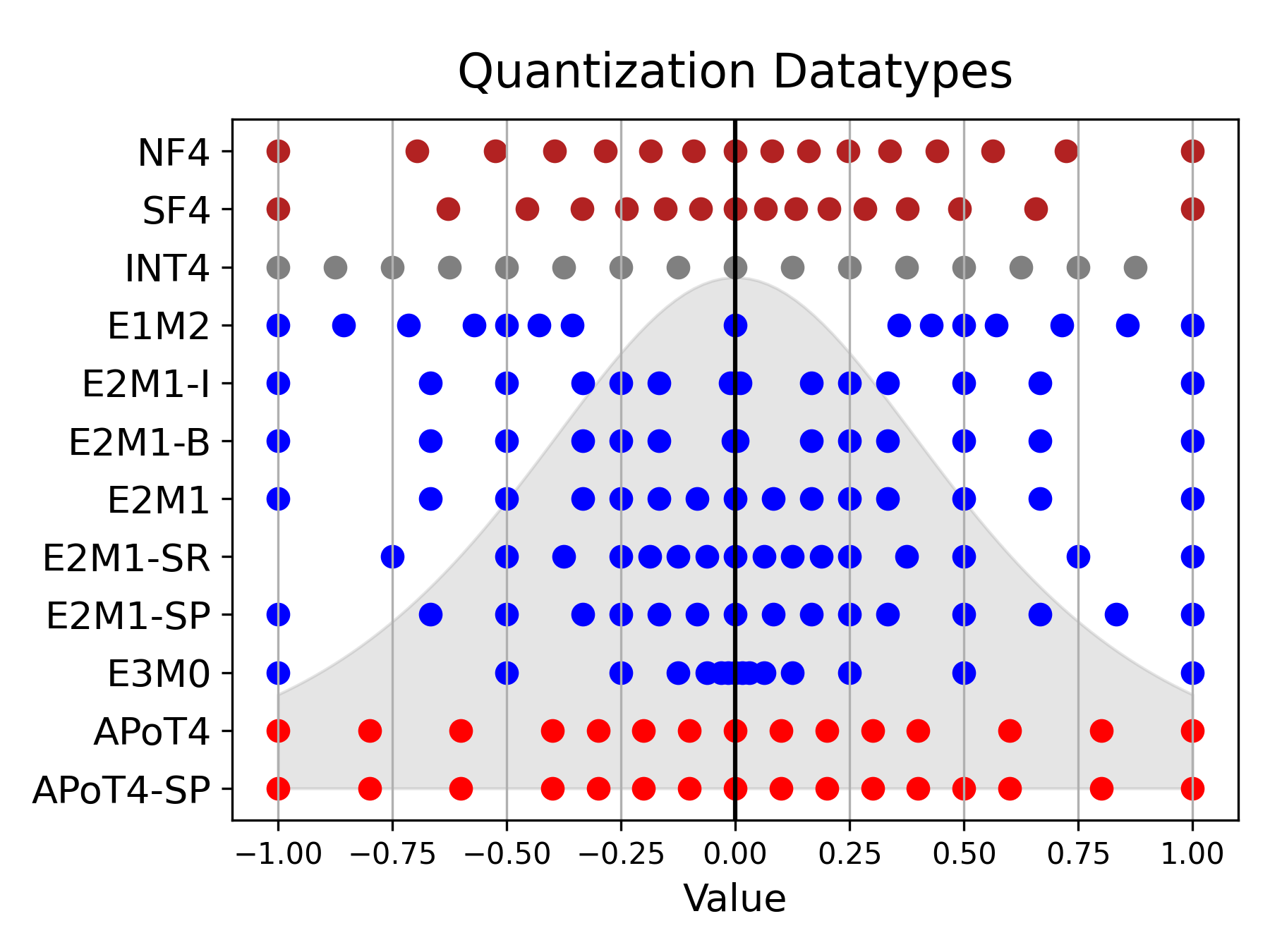}
   \vspace{-20pt}
   \caption{\textbf{Quantization Datatypes --} Datatypes should reflect LLM weight and activation distributions to achieve the highest quality. In this work, we compare model accuracy, chip area, and power consumption across datatypes to map the quality-efficiency Pareto frontier. We also propose alternative datatypes including Student Float (SF4), super-range E2M1 (SR), and super-precision E2M1 (SP). These complement existing datatypes, e.g., Normal Float (NF4), Intel E2M1 (E2M1-I), bitsandbytes E2M1 (E2M1-B) and Additive Powers of Two (APoT4).}
   \vspace{-10pt}
   \label{fig:datatypes}
\end{figure}

Quantization has become the mainstream method for deep neural network (DNN) compression~\cite{hao2021edgeai}.
Compared to alternatives like pruning, it retains original model quality at higher compression ratios~\cite{kuzmin2023pruning}, and importantly it can be applied post-training, often without any fine-tuning.
This makes it suitable for large language models (LLMs), which require significant resources during fine-tuning for gradient and optimizer state buffers.
Recent LLM quantization works have successfully lowered weight and activation precision to eight bits~\cite{frantar2023gptq, xiao2023smoothquant} and four bits~\cite{zhao2023atom, liu2023qllm, shao2023omniquant} with minimal accuracy loss.

At four bits, prior LLM quantization has focused on integer datatypes since they are supported in current DNN accelerators~\cite{jouppi2023tpu}.
However, recent work has shown eight-bit floating-point (FP8), e.g. E4M3, achieves higher accuracy compared to INT8, where E represents the exponent bits and M the mantissa bits~\cite{kuzmin2022fp8, micikevicius2022fp8}.
These improvements motivate the further study of four-bit non-integer formats, such as FP4, that can be included in next-generation accelerators.

Many of these formats are illustrated in Figure~\ref{fig:datatypes}, which includes seven FP4 variants in blue along with INT4 and multiple alternative formats.
All formats are normalized to one for comparison and placed against an example weight distribution in the background. 
Visualizing both the datatype and underlying weight distribution is important since their agreement leads to high-accuracy post-training quantization.
For example, E2M1 typically achieves higher accuracy than INT4 because it allocates more coverage to the majority of values in the center of the distribution.
This difference between datatypes is particularly important at four bits, where there are only sixteen possible values.
At higher bitwidths, most reasonable datatypes provide dense coverage across the distribution.

In addition to preserving accuracy, datatypes must have efficient multiply-and-accumulate (MAC) units, which perform nearly all of the compute-intensive LLM operations.
For instance, while E2M1 has higher accuracy, up to a 7.13\% LAMBADA improvement on Phi-2, INT4 has an 8\% smaller and more power-efficient MAC unit.
In this work, we explore this accuracy-efficiency frontier across datatypes and summarize our contributions as follows:
\begin{enumerate}
\vspace{-10pt}
    \item Conduct a large-scale profiling of the weights and activations across 30 DNNs and discover that most DNN distributions are best approximated by the Student's t-distribution.
\vspace{-5pt}
    \item Derive a theoretically optimal datatype with respect to this distribution, Student Float (SF4), and empirically verify that it improves the state-of-the-art for lookup-based quantization.
\vspace{-5pt}
    \item Propose two variants of \textit{supernormal} support for E2M1 and Additive Powers-of-Two (APoT) datatypes, using SF4 as a high-accuracy reference.
\vspace{-5pt}
    \item Plot the Pareto frontier for accuracy and performance across datatypes, comparing FP4 vs. INT4, discussing FP4 variants, and improving the accuracy of E2M1 and APoT4 with supernormal support.
\end{enumerate}

\section{Related Work}
\label{sec:related}

DNN quantization can be broadly categorized into two branches: quantization-aware training (QAT)~\cite{zhang2023binarized} and post-training quantization (PTQ)~\cite{zhao2019ocs, zhao2019hadanet, chee2024quip}.
PTQ directly performs quantization after the model has finished training, often without any training or calibration data~\cite{cai2020zeroq, nagel2019datafree}.
This approach simplifies the model quantization process but leads to lower model accuracy, especially at extremely low precision.
In this scenario, the choice of datatype is particularly important for preserving high model accuracy.
Traditionally, integer formats were the only option at low bitwidths, yet recent work has proposed new floating-point, lookup-based, and alternative formats.
At four bits, these datatypes have complex quality and performance trade-offs that affect the model accuracy, chip area, and estimated power. 

\subsection{Floating-Point}
\label{sec:floating-point}

Floating-point formats have been essential for deep learning given their ability to represent a wide range of values necessary for weights, activations, and gradients.
Recently, the Open Compute Project proposed a standard for lower-precision formats, including FP4, FP6, and \textit{micro-scaling} formats~\cite{rouhani2023microscaling}.
This standard follows prior research like VS-Quant~\cite{dai2021vsquant} and micro-exponents (MX)~\cite{rouhani2023shared}, which share scales per block and introduce multi-level scale factors.
In addition, the quantization library ``bitsandbytes"~\cite{dettmers2022llmint8} has implemented an FP4 datatype for weight-only LLM quantization.
Similarly, Intel's neural compressor, which has become a popular library for LLM compression research, offers an FP4 implementation for weight-only LLM quantization~\cite{shen2023efficient}.

In addition, multiple recent works have compared floating-point and integer formats and explored mixed-format networks~\cite{chen2023m4bram}. 
For instance, FLIQS~\cite{dotzel2023fliqs} and MoFQ~\cite{zhang2023integer} discovered that floating-point formats produce higher accuracies across vision, language, and recommendation tasks, where the differences are larger at lower precisions.
Our work continues this line of research by comparing seven different FP4 candidates across LLMs, proposing supernormal extensions to them, and mapping their quality and hardware efficiency tradeoffs.

\subsection{Logarithmic Datatypes}
\label{sec:related-logarithmic}

As floating-point formats allocate all of their bits to the exponent, they become logarithmic formats.
In this process, these formats replace costly digital multiplications with pure exponent addition~\cite{alsuhli2023number}, yet they poorly fit natural DNN distributions.
As shown in Figure~\ref{fig:datatypes}, they cluster too many values in the center of the distribution while leaving sparse coverage at the extremes.
To address this, Additive Powers-of-Two (APoT) adds two logarithmic numbers together to better match these data distributions and increase model accuracy~\cite{li2020apot}.
At four bits, APoT has the general form: $(-1)^{S}\,(2^{E} + 2^{\tilde{E}})$, where $E$ and $\tilde{E}$ are sets of powers of two. 
This leads to a potentially large search space that we explore in Appendix~\ref{sec:apot}, yet at four bits, the only reasonable variant has $E \in \{0, 2^{-1}, 2^{-2}, 2^{-4}\}$ and $ \tilde{E} \in \{0, 2^{-3}\}$.
Therefore, we focus on this variant only.
Our work maps the quality-efficiency frontier of these formats, describes the limitations of native E3M0, and introduces two variants of APoT that achieve higher accuracy with minor area overhead.

\subsection{Normal Float}
\label{sec:normal}

\begin{figure}[t]
  \centering
    \includegraphics[width=.95\linewidth]{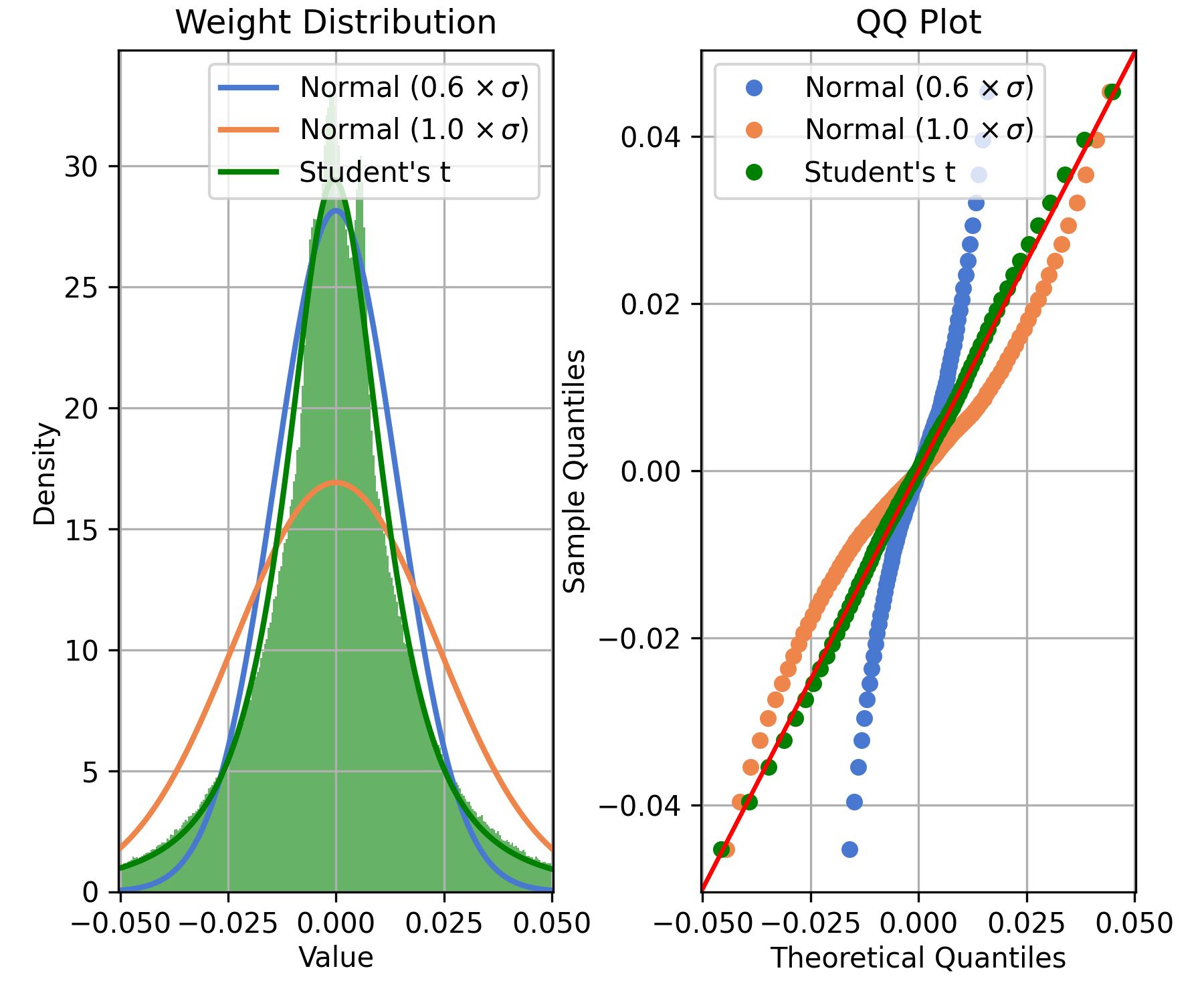}
   \vspace{-15pt}
   \caption{\textbf{Mistral-7B Weight Profile -- } The weights in Mistral-7B are best approximated by t-distributions. The best fitting normal distribution (1.0 $\times$ $\sigma$) poorly fits the peak of the distribution, and forcing it to fit the peak (0.6 $\times$ $\sigma$) causes poor representation on the larger values. Straight lines on quantile-quantile (Q-Q) plots indicate perfect fits between theoretical and sampled distributions.}
   \label{fig:profiled}
    \vspace{-15pt}
\end{figure}

While logarithmic datatypes were developed primarily for performance, Normal Float (NF4) was designed exclusively for model accuracy~\cite{dettmers2023qlora}.
It equally divides the probability mass for normal distributions using quantile functions~\cite{dettmers2022quantile}, ensuring approximately the same number of weights get mapped to each datatype value.
This leads to high accuracy, yet it relies on floating-point lookup tables and high-precision MAC units to be implemented in real hardware.
In our work, we propose an alternate lookup format, Student Float (SF4), to increase the accuracy of lookup-based quantized LLMs and build various hardware-efficient datatypes based on its insights.


\section{Proposed Datatypes}
\label{sec:formats}

\begin{table}[t]
\centering
\footnotesize
\setlength{\tabcolsep}{2pt}
\begin{tabular}{@{}rrrrrr@{}}
\toprule
\textbf{Model} & \multicolumn{2}{c}{\textbf{Weight}} & \multicolumn{2}{c}{\textbf{Activation}} \\
               & \(\nu\) & KS-\(\Delta\) & \(\nu\) & KS-\(\Delta\) \\
\midrule
OPT-1B         & 6.68\(_{2.86}\)  & 0.040 & 5.91\(_{4.08}\) & 0.117 \\
BLOOM-560M     & 5.87\(_{2.68}\)  & 0.020 & 6.75\(_{4.84}\) & 0.066 \\
BLOOM-7B       & 10.13\(_{5.96}\) & -0.019 & 4.51\(_{1.33}\) & 0.049 \\
Falcon-7B      & 5.87\(_{2.68}\)  & 0.020 & 6.75\(_{4.84}\) & 0.066 \\
LLaMA2-7B      & 6.78\(_{3.45}\)  & 0.025 & 2.98\(_{0.89}\) & 0.022 \\
Yi-6B          & 7.26\(_{4.98}\)  & 0.013 & 2.50\(_{3.30}\) & 0.036 \\
FLAN-T5        & 13.47\(_{2.40}\) & 0.004 & 5.34\(_{1.53}\) & 0.031 \\
Mistral-7B     & 1.66\(_{0.67}\)  & 0.049 & 1.67\(_{2.15}\) & 0.111 \\
Zephyr-3B      & 4.59\(_{5.20}\)  & 0.099 & 2.37\(_{1.03}\) & 0.098 \\
\midrule
BERT           & 13.13\(_{2.42}\) & -0.069 &  6.45\(_{4.35}\) & 0.034 \\
RoBERTa        & 7.28\(_{2.18}\)  & 0.022 & 6.69\(_{4.77}\) & 0.022 \\
ALBERT         & 10.87\(_{4.86}\) & 0.000 &  7.81\(_{1.75}\) & 0.018 \\
\midrule
ResNet18       & 2.71\(_{0.69}\)  & 0.069 & 10.94\(_{6.20}\) & -0.008 \\
ResNet50       & 2.95\(_{1.22}\)  & 0.052 & 6.57\(_{7.03}\) & 0.006 \\
MobileNetV2    & 5.02\(_{5.55}\)  & 0.003 & 8.22\(_{7.92}\) &  0.003 \\
EfficientNet-B0 & 4.29\(_{5.42}\) & 0.065 & 3.51\(_{1.86}\) & 0.029 \\
\bottomrule
\end{tabular}
\caption{\textbf{Weight and Activation Profiling --} DNN distributions are better approximated by t-distributions, typically with single-digit degrees of freedom (\(\nu\)). The mean and variance for \(\nu\) are calculated across layers. The Kolmogorov-Smirnov (KS) $\Delta$ measures the difference between the KS distance on the best-fit normal and Student's t-distributions. Positive values indicate a smaller distance to the t-distribution.}
\vspace{-5pt}
\label{tab:profiling}
\end{table}

In this section, we conduct a large-scale profiling of LLM weight and activation distributions across models and applications.
We then use these distributions to analytically derive the SF4 datatype and introduce supernormal support, which increases model accuracy for E2M1 and APoT4 formats with low hardware overhead.

\subsection{Student's t-Distribution}
\label{sec:student}

Instead of the normal distribution, we use the Student's t-distribution to model LLM weights and activations.
This distribution, $S(t; \nu)$, generalizes the normal distribution by introducing a degree of freedom parameter $\nu$ that controls the shapes of its peaks and tails.
Larger $\nu$ leads to wider peaks and thinner tails (shown in Appendix~\ref{sec:sf4-appendix}).
The t-distribution probability density function (PDF) is shown below, where $\Gamma$ is the generalized factorial.

\vspace{-15pt}
\begin{align}
S(t; \nu) &= \frac{\Gamma\left(\frac{\nu + 1}{2}\right)}{\sqrt{\nu\pi}\, \Gamma\left(\frac{\nu}{2}\right)} \left(1 + \frac{t^2}{\nu} \right)^{-\frac{\nu + 1}{2}}
\end{align}
\vspace{-15pt}

As $\nu \to \infty$, this distribution converges to the standard normal distribution:

\vspace{-15pt}
\begin{align}
S(t; \nu \to \infty) &= \frac{1}{\sqrt{2\pi}} e^{-\frac{t^2}{2}} 
\end{align}
\vspace{-15pt}

This distribution is useful for studying LLM weights and activations, since it can both quantify the normality of the distribution through $\nu$ and offer an analytical model for deriving future datatypes.

\subsection{Model Profiling}
\label{sec:model_profiling}

Figure~\ref{fig:profiled} (left) applies this analysis to an MLP weight tensor from Mistral-7B~\cite{jiang2023mistral}.
It shows the weight histogram along with the t-distribution and standard normal distribution.
It reveals that the best-fit t-distribution gives a better representation compared to the best-fit normal distribution ($1.0 \times \sigma$) at small and large values.
Furthermore, it shows that this is not just a matter of incorrect scaling.
Since when $\sigma$ is scaled down by 0.6 in the normal distribution to fit the peak, the larger values are no longer well-represented.
The right figure shows the same results in a quantile-quantile (Q-Q) plot, which compares the theoretical quantiles of each distribution to the profiled quantiles of the weight tensor.
In a Q-Q plot, straight lines represent perfect matches between the profiled data and theory, and therefore the t-distribution represents a significantly stronger fit overall.

\begin{algorithm}[t]
\caption{Student Float Derivation}
\begin{algorithmic}[1]
\STATE Set $\delta = \frac{1}{2}\left(\frac{1}{32} + \frac{1}{30}\right)$.

\STATE Compute eight evenly spaced probabilities $p_1, \ldots, p_8$ where $p_1 = \delta$ and $p_8 = \frac{1}{2}$, and then compute eight evenly spaced probability values $p_8, \ldots, p_{16}$ such that $p_8 = \frac{1}{2}$ and $p_{16} = 1 - \delta$.

\STATE Set $\tilde{s}_i = Q_S(p_i; \nu)$ where $Q_S$ is the quantile function for the Student's t-distribution $S(t; \nu$) with degrees of freedom $\nu$.

\STATE Normalize $\tilde{s}$ to $[-1, 1]$: $s_i = \frac{\tilde{s}_i}{\max_i |\tilde{s}_i|}$.
\label{alg:student}
\end{algorithmic}
\end{algorithm}

Table~\ref{tab:profiling} expands this analysis by quantifying the mean and variance for $\nu$ across layers in LLMs, BERT-like models, and CNNs.
It shows that the best fitting t-distributions typically have small single-digit degrees of freedom ($\nu$), with a few exceptions like the weights in FLAN-T5~\cite{wei2022finetuned}.
Since t-distributions approach normal distributions at high $\nu$, this implies they are significantly different from normal distributions.
The table also lists the difference $\Delta$ between the Kolmogorov-Smirnov (KS) distances for the best-fitting t-distribution and normal distributions.
The positive differences in most models indicate that the t-distribution has an overall better fit, and these differences also suggest that $\nu=10$ is approximately the cutoff for normal distributions.
More networks and more detailed analysis are located in Appendix~\ref{sec:profiling}.

\subsection{Student Float}
\label{sec:student-float}

Given these results, we can generate datatype optimized for the Student's t-distribution, which we refer to as Student Float (or SF4 at four bits).
In this derivation, we follow prior work~\cite{dettmers2023qlora} and equalize the expected number inputs (weights or activations) mapped to each datatype value.
This effectively equalizes the load across the datatype and ensures the quantized histogram will be approximately balanced and flat.

This process is described in Algorithm~\ref{alg:student}, which was adapted from the derivation of the NF4 datatype~\cite{dettmers2023qlora}.
It first produces sixteen numbers, $p_i$, equally spread out in probability space, although it fixes $p_8 = \frac{1}{2}$ to force a lossless representation for zero. 
This is important since quantization error on zero inputs can lead to multiple practical issues, e.g., incorrect masking or zero padding, that would need to be handled specially in software.
Additionally, it adds more values on the positive side, against the convention for integer types, since modern activation functions often bias activations toward positive values.

It then maps $p_i$ through the Student's t-distribution quantile function, $Q(p)$, to produce the unnormalized datatype values $\Tilde{s}_i$
This quantile function gives the value $x = Q(p)$, such that $S(X \leq x) = p$, where $X$ is a random variable following the t-distribution $S$.
Therefore, equally spread probabilities will be mapped to quantiles that equally divide the probability mass.
The values are finally normalized into $[-1, 1]$ for simplicity, but the true range of the datatype will be determined by the high-precision quantization scale factors at the row or group level.

\subsection{Accuracy Study}
\label{sec:sf4-accuracy}

\setlength{\tabcolsep}{2pt}
\begin{table}[t]
\centering
\footnotesize 
\begin{tabular}{rrrrrrrrrr}
\toprule
 & & \multicolumn{2}{c}{\textbf{OPT-125M}} & \multicolumn{2}{c}{\textbf{OPT-1B}} & \multicolumn{2}{c}{\textbf{Phi-2}} & \multicolumn{2}{c}{\textbf{LLaMA2-7B}} \\
\midrule
 & $\nu$ & PPL & ACC & PPL & ACC & PPL & ACC & PPL & ACC \\
\midrule
FP32 & - & 26.02 & 37.90 & 6.64 & 57.89 & 5.52 & 62.57 & 3.40 & 73.92 \\
\midrule
NF4 & - & 33.77 & 34.06 & 7.21 & 56.43 & 6.47 & 60.94 & 3.71 & 71.98 \\
\midrule
SF4 & 3 & 29.24 & 37.18 & 7.65 & 54.92 & 6.38 & 61.07 & 3.58 & 72.38 \\
SF4 & 4 & 27.21 & 37.30 & 6.95 & 57.50 & \textbf{6.26} & 61.19 & \textbf{3.52} & \textbf{72.54} \\
SF4 & 5 & \textbf{25.69} & \textbf{38.56} & 6.90 & 57.83 & 6.33 & \textbf{61.56} & 3.60 & 72.42 \\
SF4 & 6 & 25.80 & 37.90 & \textbf{6.70} & \textbf{58.59} & 6.34 & 60.92 & 3.69 & 71.82 \\
SF4 & 7 & 29.22 & 36.43 & 6.81 & 58.08 & 6.48 & 60.33 & 3.69 & 71.80 \\

\bottomrule
\end{tabular}
\caption{\textbf{Degrees of Freedom --} LLM evaluation on LAMBADA accuracy (ACC) and perplexity (PPL). SF4 achieves its highest quality when generated with the most common degrees of freedom (\(\nu\)) profiled in Table~\ref{tab:profiling}. SF4 converges to NF4 in the limit (shown in Appendix~\ref{sec:sf4-appendix}), yet its accuracy peaks around \(\nu = 5\).}
\label{tab:student}
\vspace{-5pt}
\end{table}

Given the parameterization of the quantile function, $Q_S(p; \nu)$, SF4 would vary with the choice of $\nu$. 
As $\nu$ increases, the peaks of the t-distribution become shorter and wider, SF4 spreads out more, and in the limit, it converges to NF4 (shown in Appendix~\ref{sec:sf4-appendix}).
However, since SF4 will be a reference for non-lookup datatypes with specialized and efficient MAC units, it should have a definite form and $\nu$ should be fixed across models.
Therefore, we use the most common degrees of freedom in Table~\ref{tab:profiling} and fix $\nu = 5$.

To evaluate this choice, Table~\ref{tab:student} sweeps the degrees of freedom and measures the LAMBADA accuracy and perplexity on OPT-125M, OPT-1B, Phi-2, and LLaMA2-7B.
It shows the highest accuracy and lowest perplexity results typically cluster around $\nu = 5$, although there is some variance across models.
In this table, SF4 reaches its highest accuracy significantly before converging to NF4, which occurs approximately at $\nu=10$ as discussed in Section~\ref{sec:model_profiling}.

\subsection{Supernormal Support}
\label{sec:supernormal}

\setlength{\tabcolsep}{3.5pt} 
\renewcommand{\arraystretch}{0.8} 
\begin{table*}[t!]
\centering
\footnotesize
\begin{tabular}{rrrrrrrrrrrrrrrr}
\toprule
 & & \multicolumn{2}{c}{\textbf{Mistral-7B}} & \multicolumn{2}{c}{\textbf{OPT-1B}} & \multicolumn{2}{c}{\textbf{OPT-6.7B}} & \multicolumn{2}{c}{\textbf{LLaMA2-7B}} & \multicolumn{2}{c}{\textbf{Phi-2}} & \multicolumn{2}{c}{\textbf{BLOOM-7B}} & \multicolumn{2}{c}{\textbf{Yi-6B}}\\
\midrule
Calib. Method &  & None & MSE & None & MSE & None & MSE & None & MSE & None & MSE & None & MSE & None & MSE\\
\midrule
\multirow{9}{*}{LAMBADA ↑} & FP32 & 75.90 & 75.90 & 57.89 & 57.89 &  67.69 & 67.69 & 73.92 & 73.92 & 62.57 & 62.57 & 57.64 & 57.64 & 68.27 & 68.27\\
\cmidrule(lr){2-16}
& NF4 & 74.97 & 74.97 & 56.43 & 56.37 & 67.88 & \textbf{68.43} & 71.20 & 71.98 & \textbf{61.28} & 60.94 & 57.03 & 57.09 & 67.46 & \textbf{68.19}\\
& SF4 & \textbf{75.90} & \textbf{75.00} & \textbf{58.02} & \textbf{57.83} & \textbf{68.02} & 68.02 & \textbf{71.96} & \textbf{72.42} & 60.47 & \textbf{61.56} & \textbf{57.97} &  \textbf{57.87} & \textbf{67.84} & 68.04\\
\cmidrule(lr){2-16}
& INT4 & 73.92 &  73.74 & 55.52 & 56.96 & 63.92 & 67.07 & 72.06 & 70.19 & 58.59 & 55.11 & 56.08 & 56.14 & 64.93 &  61.75\\
\cmidrule(lr){2-16}
& E2M1-I & 74.17 & 74.36 & 56.18 & 56.53 & 67.49 & 66.02 & 71.43 & 70.72 & 58.20 & 59.15 & 55.75 & 55.82 & 64.39 & 62.12\\
& E2M1-B & 73.98 & 73.65 & 55.73 & 57.13 & 66.97 & 65.55 & 70.75 & 70.68 & 58.32 & 59.91 & 55.64 & 55.72 & 63.92 & 60.64\\
& E2M1 & 74.75 & 74.81 & \textbf{56.26} & \textbf{57.52} & \textbf{67.84} & \textbf{67.86} & \textbf{72.40} & 71.51 & 59.95 & 58.92 & 56.51 &  56.48 & 66.74 & 66.95\\
& + SR & 72.95 & 72.95 & 54.41 & 54.41 & 67.26 &  67.26 & 71.07 & 71.07 & \textbf{62.24} & \textbf{62.24} & 50.18 & 50.34 & 59.97 & 60.01\\
& + SP & \textbf{75.41} &  \textbf{74.99} & 55.85 & 57.46 & 67.24 & 67.36 & 71.65 & \textbf{71.84} & 61.73 & 60.97 & \textbf{56.86} & \textbf{56.72} & \textbf{67.38} & \textbf{67.45}\\
& E3M0 & 74.23 & 71.05 & 52.36 & 53.02 & 62.64 & 64.47 & 69.92 & 68.66 & 54.96 & 55.58 & 56.47 & 56.42 & 65.15 & 65.38\\
\cmidrule(lr){2-16}
& APoT4 & \textbf{75.41} & \textbf{73.78} & \textbf{56.22} & 54.67 & \textbf{66.08 }& 67.53 & 72.77 & 71.58 & 59.62 & 59.97 & 57.02 &  57.12 & \textbf{68.19} & 68.07\\
& + SP & 75.12 & 74.05 & 55.27 & \textbf{55.25} & 65.92 & \textbf{68.06} & \textbf{73.22} & \textbf{71.63} & \textbf{61.09 }& \textbf{61.50} & \textbf{57.13} &  \textbf{57.23} & 68.04 & \textbf{68.31}\\
\midrule
\multirow{9}{*}{WikiText-2 ↓} & FP32  & 18.01 & 18.01 & 16.41 & 16.41 & 12.28 & 12.28 & 8.79 & 8.79 & 11.05 & 11.05 & 14.71 & 14.71 & 10.21 & 10.21\\
\cmidrule(lr){2-16}
& NF4   & 19.80  & 19.36 & 17.17 & 17.13 & 12.73 & 12.75 & \textbf{9.11} & 9.12 & 11.89 & 11.89 & \textbf{14.94} & \textbf{14.74} & 10.36 & 10.47\\
& SF4   &\textbf{19.09} & \textbf{19.34} & \textbf{ 17.11} & \textbf{17.10} & \textbf{12.67} & \textbf{12.66} & 9.16  &\textbf{ 9.10} & \textbf{11.83} & \textbf{11.84} & 14.96 & 14.84 & \textbf{10.34} & \textbf{10.36}\\
\cmidrule(lr){2-16}
& INT4  & 20.17 & 20.81 & 18.28 & 18.02 & 13.27 & 13.20 & 9.33  & 9.71 & 12.41 & 12.81 & 15.16 & 15.25 & 10.71 & 11.34\\
\cmidrule(lr){2-16}
& E2M1-I& 20.07 & 20.55 & 17.86  & 18.00 & 12.92 & 12.96 & 9.37 & 9.74 & 12.19 & 12.38 & 15.18 & 15.16 & 10.69 & 11.34\\
& E2M1-B& 20.93 & 21.17 & 18.34 & 18.15 & 13.11 & 13.19 & 9.43 & 9.89 & 12.37 & 12.64 & 15.22 & 15.26 & 10.76 & 11.54\\
& E2M1  & 19.76 & \textbf{19.27} & 17.24 & 17.25 & 12.78 & 12.79 & 9.17 & 9.21 & 11.97 & 11.99 & 15.01 & 15.18 & 10.42 & 10.54\\
& + SR  & 20.25 & 20.25 & 17.62 & 17.62 & 13.06 & 13.06 & 9.84 & 9.84 & 12.58 & 12.58 & 15.95 & 15.82 & 11.60 & 11.54\\
& + SP  & \textbf{19.38} & 19.47 & \textbf{17.19} & \textbf{17.18} & \textbf{12.76} & \textbf{12.77} & \textbf{9.13} &\textbf{ 9.20} & \textbf{11.92} & \textbf{11.96} & \textbf{14.98} & \textbf{14.89} & \textbf{10.37} & \textbf{10.29}\\
& E3M0  & 20.25 & 21.93 & 18.29 & 18.41 & 13.31 & 13.91 & 9.87 & 10.06 & 12.74 & 12.92 & 15.61 & 15.71 & 11.42 & 11.43\\
\cmidrule(lr){2-16}
& APoT4 & 19.13 & \textbf{19.23} & 17.47 & 17.42 & 12.84 & 12.88 & 9.15 & \textbf{9.27} & 12.09 & 12.17 & 15.02 &  14.98 & 10.46 & 10.49\\
& + SP & \textbf{18.93} & 19.32 & \textbf{17.40} & \textbf{17.32} & \textbf{12.80} & \textbf{12.85 }& \textbf{9.11} & 9.41 & \textbf{11.98} & \textbf{12.06} & \textbf{14.99} &  \textbf{14.92} & \textbf{10.40} & \textbf{10.39}\\
\bottomrule
\end{tabular}
\caption{\textbf{Weight-Only Eval --} Student Float (SF4) typically outperforms NF4, and the supernormal variants (SR and SP) often improve over E2M1 and APoT4, although there are many exceptions. All models evaluated with weight-only sub-channel quantization with block size 128 with optional MSE clipping calibration on the LAMBADA and WikiText-2 datasets.}
\label{tab:woq-lambada}
\end{table*}

Given its high accuracy, SF4 can be used as a reference for building efficient datatypes with corresponding MAC units.
Figure~\ref{fig:datatypes} visualizes five E2M1 variants next to SF4.
Assuming model accuracy is fully determined by the shape of the datatype, this figure shows the issues with multiple variants in comparison to SF4.
For example, E2M1-I and E2M1-B push their subnormal values too close to zero, which will introduce large quantization errors on the most numerous central values.

In addition, E2M1 only uses 15 unique values and SF4 uses all $2^4 = 16$ values. This missing value is caused by the floating-point sign bit, which introduces positive and negative zero.
At higher precision, such as eight bits, this redundancy wastes only 0.4\% of its bitspace, but it makes a large difference at four bits, where FP4 wastes 6.25\% of its values.
Therefore, we propose adding additional \textit{supernormal} support to E2M1 to complement the existing subnormal support.
This reassigns negative zero to a useful value and brings these formats more in line with the SF4, as shown in Figure~\ref{fig:datatypes}.
In the following sections, we evaluate the accuracy and efficiency of two supernormal variants:
\begin{enumerate}
   \vspace{-5pt}
    \item \textbf{Super-range (SR)}, which extends the range of the values by allocating one point at the edge of the distribution.
    \vspace{-5pt}
    \item \textbf{Super-precision (SP)}, which extends the precision by giving one extra value within the distribution.
    \vspace{-5pt}
\end{enumerate}

Super-precision matches the symmetry of SF4 and often achieves higher accuracy compared to super-range, yet it leads to larger chip area and power.
For instance, it decreases the WikiText-2 perplexity compared to super-range across LLMs, including LLaMA2-7B, OPT1B, and Phi-2, while increasing the area of the corresponding MAC unit by 14\%.
Finally, we also add super-precision support to the APoT4~\cite{li2020apot} datatype in an analogous way. All datatype values are listed in Appendix~\ref{sec:datatypes}.

\section{Experiments}
\label{sec:experiments}

In this section, we evaluate these proposed datatypes against previous integer, floating-point, logarithmic, and lookup-based datatypes.
These datatypes are applied with weight-only and weight-activation quantization across popular LLMs, zero-shot evaluations, and quantization methods, totaling over 4000 data points.
Finally, we show that trends found at four bits hold for lower bitwidths and prior CNN models.
The main results are shown in this section, and the remainder are listed in the Appendix.

\subsection{Weight-Only Quantization}
\label{sec:language}

\begin{table}[t]
\centering
\footnotesize
\small
\setlength{\tabcolsep}{2pt}
\begin{tabular}{@{}rrrrrrrrr@{}} 
\toprule
& \textbf{LAMB} & \textbf{Hella} & \textbf{Wino} & \textbf{PIQA} & \textbf{BoolQ} & \textbf{ARC-c} & \textbf{\(\Delta\)\%} \\ 
\midrule
FP32      & 73.92 & 57.14 & 69.14 & 78.07 & 77.74 & 43.43 & 0.00 \\
\midrule
NF4       & 72.35 & 56.55 & \textbf{69.53} & 76.99 & 77.40 & 42.49 & -1.10 \\
SF4       & \textbf{73.20} & \textbf{56.81} & 69.06 & \textbf{77.69} & \textbf{78.56} & \textbf{43.34} & \textbf{-0.22} \\
\midrule
INT4      & 72.06 & 56.53 & 69.14 & 77.31 & 76.76 & 42.92 & -1.17 \\
\midrule
E2M1-I    & 71.43 & 56.50 & 68.90 & 77.80 & 77.06 & 42.66 & -1.30 \\
E2M1-B    & 70.75 & 56.54 & 68.98 & 77.58 & 76.73 & 43.34 & -1.28 \\
E2M1      & 71.65 & 56.69 & \textbf{69.53} & 77.97 & 78.13 & 42.49 & -0.85 \\
+ SR      & 71.07 & 54.66 & 66.85 & 76.77 & 73.55 & 42.41 & -3.49 \\
+ SP      & \textbf{71.65} & \textbf{56.84} & 69.43 & \textbf{77.99} & \textbf{78.26} & \textbf{42.49} & \textbf{-0.80} \\
E3M0      & 69.92 & 54.61 & 67.64 & 76.55 & 75.32 & 39.59 & -4.32 \\
\midrule
APoT4     & 72.77 & 56.27 & 68.27 & \textbf{78.07} & 77.55 & 43.17 & -0.86 \\
+ SP      & \textbf{73.22 }& \textbf{56.56} & \textbf{68.59} & 77.69 &\textbf{ 77.68} & \textbf{43.86} & \textbf{-0.39} \\
\bottomrule
\end{tabular}%
\vspace{-5pt}
\caption{\textbf{LLaMA2-7B Weight-Only --} Accuracy improvements with SF4 and super-precision formats continue common zero-shot benchmarks. \(\Delta\)\% represents the mean relative percentage change in accuracy from FP32. All models shown in Appendix~\ref{sec:additional-tables}.}
\label{tab:woq-llama}
\end{table}

Given the memory-bound nature of LLM inference, we begin the format evaluation with weight-only quantization.
Table~\ref{tab:woq-lambada} compares all datatypes in terms of their LAMBADA~\cite{kazemi2023lambada} accuracy and WikiText-2 perplexity on weight-only quantized models. 
These metrics were chosen first because they are the most sensitive to model perturbations.
The evaluated models include Mistral-7B~\cite{jiang2023mistral}, LLaMA2-7B~\cite{touvron2023llama}, OPT-1B~\cite{zhang2022opt}, OPT-6.7B, Phi-2~\cite{li2023textbooks}, BLOOM-7B~\cite{workshop2023bloom}, and Yi-6B.

The models were quantized and evaluated with a modified version of the neural compressor library, which includes lookup-based quantization for the new datatypes.
All models use symmetric, sub-channel quantization with block size 128, with either no clipping or weight-based MSE clipping.
This block size was selected since it is small enough to significantly increase model accuracy but large enough to align most MAC units without requiring splitting accumulations. 
Both clipping methods were included to ensure the datatype accuracy was not heavily dependent on the quantization algorithm itself.

This table demonstrates that SF4 improves model quality compared to NF4 in most cases.
In addition, it shows the FP4 variants, even in the worst case, typically outperform INT4, which agrees with the results seen in prior higher-precision comparisons to integer formats~\cite{dotzel2023fliqs, kuzmin2022fp8}.
Within these FP4 formats, the Intel and bitsandbytes variants consistently underperform compared to the E2M1 baseline, which is due to their concentrated subnormal values shown in Figure~\ref{fig:datatypes}.
Finally, the baseline APoT datatype often performs well against E2M1 and INT4, for example, increasing LAMBADA accuracy loss by 1.44\% compared to INT4 on LLaMA2-7B.
Table~\ref{tab:woq-lambada} further shows that supernormal support typically increases model quality, yet there are instances when the baseline format achieves higher accuracy.

\subsection{Zero-Shot Evaluation}
\label{sec:woq-zero}

While LAMBADA and WikiText-2 are the most sensitive metrics, other zero-shot evaluations align more closely with real-world LLM usage.
Table~\ref{tab:woq-llama} expands the weight-only comparison to include multiple zero-shot tasks evaluated on LLaMA2-7B.
It includes common-sense reasoning with HellaSwag~\cite{zellers2019hellaswag} and language comprehension with WinoGrande~\cite{sakaguchi2019winogrande} and BoolQ~\cite{clark2019boolq}.
In addition, it measures physical commonsense with PIQA~\cite{bisk2019piqa} and scientific question-answering with ARC-c~\cite{moskvichev2023conceptarc}.
Its results reinforce the previous observations, showing consistent improvements of SF4 over NF4 and improvement of the super-precision variants of E2M1 and APoT4 over their baselines.
For instance, SF4 improves over NF4 by close to 1\% on LAMBADA, PIQA, BoolQ, and ARC-c, and the inclusion of super-precision reduces accuracy loss by 0.47\% with APoT4.

\subsection{Subchannel Sweep}
\label{sec:woq-subchannel}

\begin{table}[t]
\centering
\footnotesize
\small
\setlength{\tabcolsep}{2pt}
\begin{tabular}{@{}rrrrrrr@{}} 
\toprule
\textbf{Block Size} & \textbf{16} & \textbf{32} & \textbf{64} & \textbf{128} & \textbf{256} & CW \\
\midrule
NF4       & -1.19 & \textbf{-0.89} & -1.79 & -1.87 & \textbf{-1.44} & -4.86 \\
SF4  & \textbf{-1.04} & -1.04 & \textbf{-1.38} & \textbf{-1.33 }& \textbf{-1.44} & \textbf{-3.69} \\
\midrule
INT4      & -1.98 & -2.27 & -2.27 & -2.96 & -3.53 & -7.98 \\
\midrule
E2M1-I    & -1.90 & -1.70 & -2.02 & -2.67 & -3.37 & -6.57 \\
E2M1-B    & -2.33 & -2.00 & -2.17 & -2.80 & -3.90 & -8.58 \\
E2M1      & -1.27 & -1.59 & -1.67 & -1.40 & -1.62 & -3.92 \\
+ SR      & -13.54 & -4.98 & -1.91 & -1.86 & -1.58 & \textbf{-3.21} \\
+ SP      & \textbf{-0.39} & \textbf{-0.97} & \textbf{-0.92} & \textbf{-0.66} & \textbf{-0.92 }& -3.85 \\
E3M0      & -3.25 & -3.33 & -4.20 & -4.50 & -5.77 & -6.17 \\
\midrule
APoT4      & -1.34 & -2.04 & -2.34 & -1.90 & -2.30 & -4.35 \\
+ SP      & \textbf{-0.64} & \textbf{-1.47} & \textbf{-1.13} & \textbf{-1.29} & \textbf{-1.64} & \textbf{-3.43} \\
\bottomrule
\end{tabular}%
\vspace{-5pt}
\caption{\textbf{Phi-2 Subchannel Sweep} -- Differences between formats exist even with the smallest subchannel block sizes. All results are from Phi-2 with weight-only quantization. The average relative accuracy change (↑) from FP32 is shown, calculated across LAMBADA, HellaSwag, Winogrande, PIQA, BoolQ and ARC-c. More positive change, i.e., less accuracy drop, is preferred. Channelwise (CW) quantization is shown in the last column. }
\label{tab:subchannel}
\end{table}

Subchannel quantization is standard for weight-only LLM quantization, yet the size of the subchannels affect the shape of the weight distribution and potentially the optimal format.
Therefore, Table~\ref{tab:subchannel} compares formats on Phi-2 while varying subchannel block size.
It aggregates all metrics into a single score that measures the average relative accuracy drop from FP32.
As expected, decreasing block size leads to higher accuracy across formats, yet the differences between formats still remain.
Even at the extreme with block size 16, which is beyond what current DNN accelerators can support efficiently, the trends hold from previous evaluations.
For instance, without subchannel quantization, the difference between INT4 and E2M1-SP is 4.14\% on average, and with block size 16 the difference remains at 1.59\%. 

\subsection{GPTQ Comparison}
\label{sec:GPTQ}

\begin{table}[t]
\centering
\footnotesize
\small
\setlength{\tabcolsep}{2pt}
\begin{tabular}{@{}rrrrr@{}} 
\toprule
 & \multicolumn{2}{c}{\textbf{Channelwise}} & \multicolumn{2}{c}{\textbf{Subchannel}} \\
 \addlinespace

& \textbf{RTN} & \textbf{GPTQ} & \textbf{RTN} & \textbf{GPTQ} \\
\midrule
NF4         & -4.86 & \textbf{-2.48} & -1.87 & \textbf{-1.14} \\
SF4         & \textbf{-3.69} & -2.49 & \textbf{-1.33} & -1.65 \\
\midrule
INT4        & -7.98 & -6.45 & -2.96 & -2.39 \\
\midrule
E2M1-I      & -6.57 & -5.47 & -2.67 & -2.31 \\
E2M1-B      & -8.58 & -5.35 & -2.80 & -2.46 \\
E2M1        & -3.92 & -2.57 & -1.40 & -1.48 \\
+ SR        & \textbf{-3.21} & \textbf{-2.19} & -1.86 & \textbf{-1.17} \\
+ SP        & -3.85 & -2.35 & \textbf{-0.66} & -1.54 \\
E3M0        & -6.17 & -4.76 & -4.50 & -3.64 \\
\midrule
APoT4        & -4.35 & -3.80 & -1.90 & -1.89 \\
+ SP        & \textbf{-3.43} & \textbf{-2.91} & \textbf{-1.29} & \textbf{-1.46} \\
\bottomrule
\end{tabular}%
\vspace{-5pt}
\caption{\textbf{Phi-2 GPTQ} -- Quality differences remain with weight-only quantization with the inclusion of GPTQ. The average accuracy drop (\%) is shown, calculated across LAMBADA, HellaSwag, Winogrande, PIQA, BoolQ, and ARC-c. Round-to-nearest (RTN) quantization is the baseline and results are evaluated with and without subchannel quantization with 128-element subchannels.}
\label{tab:phi-gptq}
\vspace{-15pt}
\end{table}

In addition to extreme subchannel quantization, we evaluate the effects of advanced post-training quantization like  GPTQ~\cite{frantar2023gptq}.
GPTQ is a popular weight-only optimizer that uses second-order Hessian information to improve quantization quality by iteratively updating unquantized weight blocks to account for the add quantization error. 
These results are shown in Table~\ref{tab:phi-gptq} evaluated on the Phi-2 model, where GPTQ typically reduces the accuracy loss across datatypes with and without subchannel quantization. 
However, the differences between formats remain even in this more optimized regime.

\subsection{Three-Bit Formats}
\label{sec:other-bitwidths}

The lookup datatypes NF4 and SF4 can be generalized to other precisions with slight modifications to Algorithm~\ref{alg:student}.
At three bits, Table~\ref{tab:language-3} evaluates OPT-1B across a similar subset of tasks.
This table demonstrates that at lower bitwidths, Student Float continues to outperform Normal Float across most evaluations, particularly on the more sensitive LAMBADA and Wikitext-2 metrics with an improvement of 1.13\% and 2.50\% respectively.

Of the possible FP3 datatypes, only E2M0 is well-defined, and it performs better than INT3 in all cases, which is in contrast to E3M0, where INT4 typically has higher quality.
This is because at low precision, the dynamic range of the exponent is restricted, and E2M0 becomes close in shape to SF3 (shown in Appendix~\ref{sec:datatypes}).
At two bits, the datatype shape is not well-defined and therefore it is not evaluated.

\subsection{Weight-Activation Quantization}
\label{sec:wa-quantization}

\begin{table}[t]
\centering
\footnotesize
\setlength{\tabcolsep}{3pt}
\begin{tabular}{@{}rrrrrrr@{}}
\toprule
 & \textbf{LAMB ↑} & \textbf{Hella ↑} & \textbf{Wino ↑} & \textbf{PIQA ↑} & \textbf{BoolQ ↑} & \textbf{Wiki ↓} \\
\midrule
FP32          & 57.89 & 41.54 & 59.51 & 71.71 & 57.83 & 16.41 \\
\midrule
NF3           & 46.28 & \textbf{38.10} & 54.93 & 68.06 & 53.01 & 25.06 \\
SF3           & \textbf{47.41} & 36.90 & \textbf{56.99} & \textbf{68.82} & \textbf{53.27} & \textbf{22.56} \\
\midrule
INT3          & 00.97 & 27.66 & 49.96 & 56.37 & 40.34 & 33.12 \\
\midrule
E2M0          & 23.52 & 32.43 & 53.99 & 64.15 & 51.96 & 28.98 \\
\bottomrule
\end{tabular}%
\vspace{-5pt}
\caption{\textbf{Three-Bit OPT-1B --} The same procedures for generating SF4 and NF4 can be applied at lower bitwidths. Student Float continues to improve over Normal Float in most cases, and both achieve higher accuracy than integer and floating point. }
\label{tab:language-3}
\end{table}

\begin{table}[t]
\centering
\footnotesize
\small
\setlength{\tabcolsep}{1.5pt}
\begin{tabular}{@{}crrrrrrrr@{}} 
\toprule
& & \textbf{M-7B} & \textbf{O-1B} & \textbf{O-6B} & \textbf{L-7B} & \textbf{P-2B} & \textbf{B-7B} & \textbf{Y-6B} \\ 
\midrule
\multirow{10}{*}{\raisebox{-1.5cm}{\hspace{-1em}\rotatebox[origin=c]{90}{No SmoothQuant}}} & NF4      & -4.49 & -11.02 & \textbf{-4.27} & \textbf{-2.65} & -8.00 & -8.50 & -10.61 \\
& SF4      & \textbf{-3.98} & \textbf{-10.95} & -4.76 & -2.82 & \textbf{-6.79} & \textbf{-7.39} & \textbf{-9.17} \\
\cmidrule{2-9}
& INT4     & -8.74 & -20.72 & -9.44 & -6.27 & -16.19 & -17.94 & -24.37 \\
\cmidrule{2-9}
& E2M1-I   & -8.46 & -16.00 & -5.62 & -6.11 & -15.66 & -12.40 & -17.97 \\
& E2M1-B   & -10.33 & -15.92 & -6.22 & -7.47 & -17.82 & -14.84 & -21.45 \\
& E2M1     & \textbf{-5.08} & -11.09 & \textbf{-4.16} & \textbf{-2.68} & -8.41 & -9.32 & -11.52 \\
& + SR     & -13.02 & -11.10 & -6.92 & -12.28 & -8.53 & -7.48 & -31.46 \\
& + SP     & -3.88 & -12.03 & -4.52 & -3.42 & \textbf{-7.25} & -8.97 & \textbf{-10.30} \\
& E3M0     & -8.40 & \textbf{-10.74} & -8.19 & -10.66 & -15.25 & \textbf{-6.20} & -10.56 \\
\cmidrule{2-9}
& APoT4    & \textbf{-5.46} & -12.78 & \textbf{-4.62} & -3.74 & -9.62 & -10.20 & \textbf{-12.59} \\
& + SP     & -5.68 & \textbf{-12.02} & -4.85 & \textbf{-3.50} &\textbf{ -8.48} & \textbf{-9.59} & -12.81 \\
\midrule
\midrule
\multirow{10}{*}{\raisebox{-1.5cm}{\hspace{-1em}\rotatebox[origin=c]{90}{SmoothQuant}}} & NF4      & -3.75 & \textbf{-9.66} & -1.77 & -3.60 & -6.98 & -4.49 & -5.46 \\
& SF4      & \textbf{-2.86} & -10.02 & \textbf{-1.39} & \textbf{-3.45} & \textbf{-5.86} & \textbf{-2.19} & \textbf{-3.76} \\
\cmidrule{2-9}
& INT4     & -7.09 & -10.93 & -3.60 & -6.35 & -19.97 & -11.58 & -11.52 \\
\cmidrule{2-9}
& E2M1-I   & -7.20 & -11.17 & -2.74 & -5.60 & -17.27 & -8.64 & -10.32 \\
& E2M1-B   & -7.71 & \textbf{-10.10} & -3.59 & -6.63 & -22.07 & -10.74 & -13.05 \\
& E2M1     & \textbf{-3.77} & -10.71 & -1.34 & -3.44 & \textbf{-7.57} & -4.23 & \textbf{-5.93} \\
& + SR     & -15.52 & -10.49 & -5.45 & -13.14 & -8.02 & -5.23 & -26.38 \\
& + SP     & -3.95 & -11.87 & \textbf{-1.18} & \textbf{-3.24} & -7.98 & \textbf{-4.19} & -6.24 \\
& E3M0     & -8.01 & -10.75 & -6.39 & -9.13 & -13.05 & -6.71 & -9.77 \\
\cmidrule{2-9}
& APoT4    & \textbf{-4.54} & \textbf{-9.36} & -2.10 & -4.23 & -9.82 & -6.34 & -6.40 \\
& + SP     & -4.55 & -9.76 & \textbf{-1.65} & \textbf{-4.19} & \textbf{-8.20} & \textbf{-5.63} & \textbf{-6.20} \\
\bottomrule
\end{tabular}%
\vspace{-5pt}
\caption{\textbf{W4A4 Eval --} Evaluation of W4A4 quantization averaged across LAMBADA, HellaSwag, Winogrande, PIQA, BoolQ and ARC-c. Each value represents the mean relative percentage accuracy change (↑) from FP32.}
\label{tab:qa}
\end{table}

Since MAC units require both inputs to be quantized, it is important to also evaluate weight and activation quantization.
Table~\ref{tab:qa} performs this evaluation across all the previously mentioned models and metrics, showing the average accuracy change from FP32 baseline.
Across formats, the accuracy drops are naturally larger compared to weight-only quantization, e.g. INT4 dropping 24.37\% on Yi-6B.
Yet, in many cases, the drop is limited by including SmoothQuant~\cite{xiao2023smoothquant}, which transfers the quantization difficulty from activations to weights, reducing the accuracy for INT4 to only 11.52\% on Yi-6B.

NF4 and SF4 are included in this table, even though as lookup-based datatypes, they would require custom support like product quantization to handle quantized activations~\cite{abouelhamayed2024pqa}.
Regardless of support, they are still meaningful references for designing other datatypes.
As before, these formats typically outperform the hardened datatypes, with SF4 achieving the highest overall accuracy with and without SmoothQuant, e.g. limiting the accuracy loss to an average of 2.86\% on Mistral-7B.
All of the raw table data are listed in Appendix~\ref{sec:additional-tables}.

\subsection{Vision Models}
\label{sec:vision}

\begin{table}[t]
\centering
\footnotesize
\setlength{\tabcolsep}{3pt}
\begin{tabular}{@{}rrrrr@{}}
\toprule
& \textbf{ResNet18} & \textbf{ResNet50} & \textbf{Dense121} & \textbf{ViT-B-16} \\ 
\midrule
FP32      & 69.76 & 76.13 & 74.43 & 81.07 \\
\midrule
NF4      & 58.04 & 67.66 & 68.76 & 79.48 \\
SF4      & \textbf{63.12} & \textbf{69.05} & \textbf{69.48} & \textbf{80.28} \\
\midrule
INT4     & 40.09 & 29.36 & 47.48 & 77.61 \\
\midrule
E2M1    & 55.39 & 64.47 & 67.74 & 79.66\\
+ SR    & 57.04 & 66.80 & 67.97 & 79.57 \\
+ SP    & \textbf{61.10 }& \textbf{68.31} & \textbf{68.81} & \textbf{79.94}\\
E3M0    &  49.70 & 50.04 & 53.98 & 78.99\\
\midrule
APoT4    &  54.66 & 65.13 & 62.34 & 78.96\\
+ SP    &  \textbf{55.03} & \textbf{66.09} & \textbf{63.11} & \textbf{79.04}\\

\bottomrule
\end{tabular}%
\vspace{-5pt}
\caption{\textbf{Vision Models --} Given their similar distributions, vision models have similar improvements with SF4 and super-precision formats. All models are evaluated on ImageNet using channel-wise weight and activation quantization, with clipping thresholds determined statically over 256 training examples. }
\vspace{-5pt}
\label{tab:vision}
\end{table}

Since the weights and activations for LLMs and convolutional neural networks (CNNs) follow the same distributions according to Table~\ref{tab:profiling}, we expect similar quality trends on CNNs that were found with LLMs.
Table~\ref{tab:vision} shows these results on ResNet18~\cite{he2015deep}, ResNet50, DenseNet121~\cite{huang2018densely}, and ViT-B-16 with weight and activation quantization.
SF4 again improves over NF4 and reaches the highest accuracies in all models.
For instance, it improves ResNet18 by 5.08\% when evaluated on ImageNet-1K.
Super-precision also outperforms the E2M1 and APoT4 baselines, where
E2M1 improves by up to $5.71\%$ and APoT4 by 0.96\%.

\section{Hardware Comparison}
\label{sec:performance}

In addition to maintaining high model quality, datatypes must also be efficient in real hardware.
To examine the hardware cost of different datatypes, we model their MAC units using SystemVerilog and then use Synopsys Design Compiler to synthesize their area and estimate their power under TSMC 28nm technology.
Each MAC unit contains a multiplier and an accumulator that has been sized to iteratively add 256 terms from a dot product.

\subsection{Area and Power}
\label{sec:area-power}

\begin{table}[t]
\centering
\footnotesize
\setlength{\tabcolsep}{3pt}
\begin{threeparttable}
\begin{tabular}{@{}rcrrrrr@{}}
\toprule
 & \textbf{Accum.} & \textbf{Mult.} & \textbf{Accum.} & \multicolumn{2}{c}{\textbf{MAC}} & \textbf{Rel. Chip} \\
\textbf{}      & \textbf{Bits}   & $\mu m^2$      & $\mu m^2$       & $\mu m^2$ & $\mu W$        & \textbf{Overhead \tnote{1}} \\
\midrule
INT4            & 16              & 75.3           & 85.4            & 160.7     & 48.5               &     $0.0\%$             \\
INT5            & 18              & 106.6           & 97            & 203.6     & 59.8               &     $17.7\%$             \\
\midrule
E2M1-I          & 20           & 119.1            & 109.1     & 228.2              &   59.7             & $4.2\%$ \\
E2M1-B          & 23              & 137.9           & 131           & 268.9     & 67.9            &    $6.7\%$               \\
E2M1           & 17              & 79.7           & 90.7            & 170.4     & 49.6               &   $0.6\%$               \\
+ SR           & 18              & 96.8          & 94.5            & 191.3     & 53.5               &    $1.9\%$              \\
+ SP           & 19              & 121.5          & 96.5           & 218.0     & 54.6               &     $3.6\%$             \\
E3M0           & 22              & 98.0           & 119.7           & 217.7     & 59.5               &     $3.6\%$              \\
\midrule
APoT4           & 16              & 96.2          & 85.4           & 181.6     & 47.2               &     $1.3\%$              \\
+ SP           & 16              & 99.7          & 85.4           & 185.1     & 45.5               &     $1.5\%$              \\
\bottomrule
\end{tabular}%
\begin{tablenotes}
  \item[1] Assuming the MAC units and the memory system occupy 10\% and 60\% of the chip area, respectively \cite{chen2019eyeriss, jouppi2021lessons}.
\end{tablenotes}
\end{threeparttable}
\caption{\textbf{Hardware Results --} Area and power measurements for the MAC units for each datatype. The relative system overhead represents the area overhead of each format compared to INT4, accounting for the other components of a DNN accelerator.}
\vspace{-5pt}
\label{tab:hardware}
\end{table}

Table~\ref{tab:hardware} summarizes these hardware costs across datatypes and adjusts the accumulation bitwidth for lossless accumulation in integer or fixed-point.
This assumption means that each format must vary its accumulator bitwidth to avoid overflow and underflow, which can have a significant effect on the total area.
At low precision, this accumulator area can even exceed the multiplier area, especially with format with larger dynamic range. For example, the E2M1 accumulator is 13.8\% larger than its multiplier.
This is typically not true at higher precision, since multipliers scale quadratically with bitwidth while accumulators only scale linearly.  

This table shows that, despite often having the lowest accuracy, INT4 remains the most efficient format due to its small accumulator.
Other formats, which have larger dynamic ranges, increase the required multiplier accumulator bitwidth, leading to a larger total area of the MAC unit.

However, the MAC unit is only one part of the whole system, which involves memory, communication, and additional control components.
To account for these, Table~\ref{tab:hardware} includes a column for estimated system chip overhead with respect to INT4.
This estimate assumes the MAC units and memory occupy approximately $10\%$ and $60\%$ area of the entire design, respectively, which is common within modern DNN accelerators~\cite{chen2019eyeriss, jouppi2023tpu}.
Since the memory system is largely unaffected for a given bitwidth, the increased area for compute is dampened at the system level.
For instance, while the MAC area overhead of adding super-precision support to E2M1 is 27.9\%, its overall chip area overhead is only 3.6\%.

\subsection{Higher Bitwidths}
\label{sec:higher-bitwidths}

\begin{figure}[t!]
  \centering
  \includegraphics[width=\linewidth]{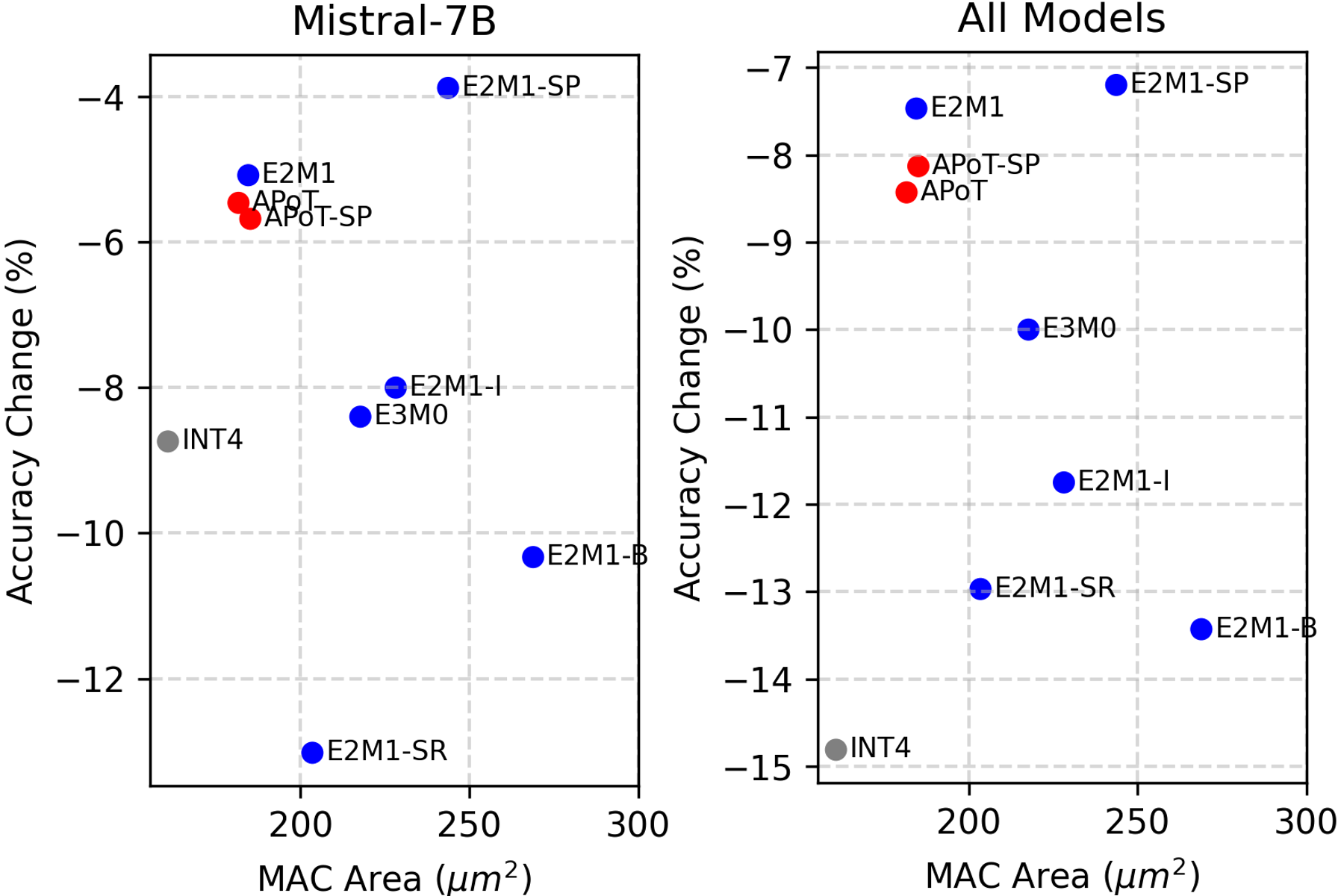}
  \vspace{-10pt}
      \caption{\textbf{Quality vs. Area --} Relative accuracy change from unquantized baselines averaged across LAMBADA, HellaSwag, Winogrande, PIQA, BoolQ and ARC-c. All Model results averaged across Mistral-7B, OPT-1B, OPT-6.7B, LLaMA2-7B, Phi-2, BLOOM-7B, and Yi-6B. All individual model Paretos are shown in Appendix~\ref{sec:additional-paretos}.}
  \vspace{-10pt}
  \label{fig:pareto}
\end{figure}

In addition to non-traditional formats, future accelerators can increase the bitwidth beyond four bits.
To consider this possibility, Table~\ref{tab:hardware} includes the estimated area and power for INT5, which would outperform all four-bit formats in model quality.
It would even achieve this with a comparable MAC area compared to some four-bit datatypes.
However, it would add significant memory overhead that leads to a large increase in the overall system area.
For example, although the MAC area of INT5 only increases by $2.7\%$ over INT4, the required memory is at least $1.25\times$ higher, leading to $17.7\%$ system overhead in total.

\subsection{Quality vs. Area}
\label{sec:accuracy-area}

Combining the quality and performance results, Figure~\ref{fig:pareto} plots the average accuracy changes across models and tasks.
It also highlights the Mistral-7B model, leaving the other models in Appendix~\ref{sec:additional-paretos}.
The accuracy change is evaluated across the same tasks in Table~\ref{tab:qa} with respect to the unquantized FP32 baseline.
This figure shows a Pareto curve from INT4 at the lowest area and quality to super-precision E2M1 with the highest area and quality.
It first demonstrates the strength of E2M1 compared to INT4, since it can significantly reduce the average accuracy drop across models by 7.34\% with a near negligible system overhead of 0.6\%.
The APoT datatypes are typically in the middle of the curve, with accuracies close to E2M1.
However, APoT requires additional logic to be converted from higher-precision FP32 or BF16, and therefore it becomes less useful than E2M1 in real systems.

In addition, super-precision offers accuracy boosts to E2M1 across models.
With approximately a 3\% system area overhead, super-precision could be worth the extra complexity as it would enable more LLM applications at four bits.
Other formats such as the Intel and bitsandbytes variants of E2M1 and E3M0 are strictly worse; they have higher dynamic range, which increases the size of the accumulator, and they nearly always reduce model accuracy compared to E2M1.

\section{Conclusion}
\label{sec:conclusion}

DNN quantization has become essential for enabling LLM applications to reach latency targets and reduce infrastructure costs.
Traditionally, these quantization methods have relied on integer datatypes, yet the recent success of FP8 formats motivates further study of non-integer formats at four bits.
In this work, we first profile over 30 DNNs and discover most have weights and activations that are best approximated by the Student's t-distribution.
Then, by optimizing for this distribution, we introduce Student Float (SF4), which can be used as a drop-in replacement for NF4 in memory-bound applications involving weight-only quantization.
We first find it increases model quality across the most popular LLMs and then use these insights to analyze more efficient datatypes.
For example, the high accuracy of E2M1 over INT4 stems from its piecewise approximation of SF4.
These high-quality datatypes reduce the need for more complex algorithmic optimizations such as SmoothQuant, GPTQ, and fine-grained subchannel quantization.
This decreases the system complexity, such as maintaining SmoothQuant scales on residual branches and optimizing low block-size subchannel quantization, and lowers the effort for high-quality LLM quantization.

Finally, we introduce supernormal extensions to E2M1 and APoT to increase their model accuracies at the cost of minor increases in system area.
We then map out the Pareto frontier across datatypes in terms of model accuracy and chip area.
This frontier begins with INT4 with lowest accuracy but highest efficiency and extends to E2M1 with super-precision with highest accuracy and close to highest area.
In particular, we find that E2M1 with supernormal support increases the accuracy of Phi-2 by up to 2.19\% with 1.22\% estimated chip overhead, offering a promising option to enable new quality-neutral LLM applications at four bits.

\section*{Impact Statement}
\label{sec:impact}

This paper presents a large comparison of formats evaluated on modern large language models and proposes multiple new formats with strong quality-efficiency tradeoffs.
There are no specific societal impacts
of this work that do not apply equally to the general LLM
literature.
\section*{Acknowledgements}
\label{sec:acknowledgements}
This work is supported in part by National Science Foundation (NSF) award \#2019306. 
We additionally acknowledge Guanlin Zhu, Solomon Lee, and Xingze Li for general discussions and technical explorations around this work. In addition, we thank Yun Ni, Andrew Li, Garrett Anderson, Cheng Fu, Yin Zhong, Ritesh Patel, and Lifeng Nai. Their insights and suggestions during multiple discussions were helpful for guiding and refining this work.

\bibliography{bib}
\bibliographystyle{icml2024}

\clearpage
\appendix
\clearpage
\setcounter{page}{1}

\section{Weight and Activation Profiling}
\label{sec:profiling}

\begin{table}[t]
\centering
\footnotesize
\setlength{\tabcolsep}{2pt}
\begin{tabular}{@{}rrrrrr@{}}
\toprule
\textbf{Model} & \multicolumn{2}{c}{\textbf{Weight}} & \multicolumn{2}{c}{\textbf{Activation}} \\
               & \(\nu\) & KS-\(\Delta\) & \(\nu\) & KS-\(\Delta\) \\
\midrule
GPT2        & 2.04\(_{0.86}\) & 0.086 & 7.21\(_{2.13}\) & 0.097 \\
OPT-1B         & 6.68\(_{2.86}\)  & 0.040 & 5.91\(_{4.08}\) & 0.117 \\
BLOOM-560M     & 5.87\(_{2.68}\)  & 0.020 & 6.75\(_{4.84}\) & 0.066 \\
BLOOM-7B       & 10.13\(_{5.96}\) & -0.019 & 4.51\(_{1.33}\) & 0.049 \\
Falcon-7B      & 5.87\(_{2.68}\)  & 0.020 & 6.75\(_{4.84}\) & 0.066 \\
LLaMA2-7B      & 6.78\(_{3.45}\)  & 0.025 & 2.98\(_{0.89}\) & 0.022 \\
Yi-6B          & 7.26\(_{4.98}\)  & 0.013 & 2.50\(_{3.30}\) & 0.036 \\
T5-Small        & 11.80\(_{4.01}\) & 0.004 & 6.74\(_{2.94}\) & 0.021 \\
FLAN-T5        & 13.47\(_{2.40}\) & 0.004 & 5.34\(_{1.53}\) & 0.031 \\
Mistral-7B     & 1.66\(_{0.67}\)  & 0.049 & 1.67\(_{2.15}\) & 0.111 \\
Zephyr-3B      & 4.59\(_{5.20}\)  & 0.099 & 2.37\(_{1.03}\) & 0.098 \\
\midrule
BERT           & 13.13\(_{2.42}\) & -0.069 &  6.45\(_{4.35}\) & 0.034 \\
RoBERTa        & 7.28\(_{2.18}\)  & 0.022 & 6.69\(_{4.77}\) & 0.022 \\
ALBERT         & 10.87\(_{4.86}\) & 0.000 &  7.81\(_{1.75}\) & 0.018 \\
\midrule
VGG19          &  5.96\(_{2.24}\) & 0.016 & 1.81\(_{0.75}\) & 0.095 \\
ResNet18       & 2.71\(_{0.69}\)  & 0.069 & 10.94\(_{6.20}\) & -0.008 \\
ResNet50       & 2.95\(_{1.22}\)  & 0.052 & 6.57\(_{7.03}\) & 0.006 \\
ResNet101       & 1.96\(_{0.84}\) & 0.075 & 9.26\(_{5.13}\) & 0.008 \\
InceptionV3       & 2.61\(_{0.83}\) & 0.044 & 12.02\(_{4.62}\) & 0.002\\
InceptionV4       & 2.29\(_{1.55}\) & 0.007 & 9.18\(_{6.11}\) & -0.039\\
MNASNet100       & 4.45\(_{4.27}\) & 0.020 & 9.84\(_{5.56}\) & 0.021\\
MobileNetV2    & 5.02\(_{5.55}\)  & 0.003 & 8.22\(_{7.92}\) &  0.003 \\
MobileNetV3    & 4.35\(_{3.16}\)  & 0.031 & 7.82\(_{5.98}\) &  0.581 \\
EfficientNet-B0 & 4.29\(_{5.42}\) & 0.065 & 3.51\(_{1.86}\) & 0.029 \\
\midrule
ConvNext-S & 1.96\(_{0.79}\) & 0.110 & 4.59\(_{4.07}\) & 0.069 \\
RegNet &  2.91\(_{1.78}\) & 0.075 & 6.12\(_{2.37}\) & 0.037 \\
ConvMixer       & 2.45\(_{1.16}\) & 0.125 & 9.84\(_{5.56}\) & 0.021\\
CoAT-Lite       & 2.11\(_{1.87}\) & 0.050 & 7.29\(_{5.28}\) & -0.006\\
PiT-B       & 8.13\(_{3.25}\) & 0.006 & 8.87\(_{4.22}\) & 0.017\\

\bottomrule
\end{tabular}
\caption{\textbf{Profiling --} DNN distributions are better approximated by t-distributions, typically with single-digit degrees of freedom (\(\nu\)). The mean and variance for \(\nu\) are calculated across layers. The Kolmogorov-Smirnov (KS) $\Delta$ measures the difference between the KS distance run on the best-fit normal and Student's t-distributions. Positive values indicate a smaller distance to the t-distribution. For activation profiling, model inputs are randomly generated.}
\vspace{-10pt}
\label{tab:profiling-additional}
\end{table}

\begin{table}[t]
\centering
\footnotesize
\setlength{\tabcolsep}{2pt}
\begin{tabular}{@{}rrrrrr@{}}
\toprule
\textbf{Model} & \multicolumn{2}{c}{\textbf{Weight}} & \multicolumn{2}{c}{\textbf{Activations}} \\
               & \(\nu\) & KS-\(\Delta\) & \(\nu\) & KS-\(\Delta\) \\
\midrule
Query       & 9.88\(_{4.78}\)  &-0.008 & 3.77\(_{0.46}\) &  0.027 \\
Key       & 9.48\(_{4.85}\)  &  -0.001 & 11.07\(_{4.56}\) & -0.002 \\
Value       & 13.83\(_{2.10}\)  & -0.001 & 9.40\(_{4.33}\) & 0.002 \\
Out       & 8.77\(_{4.50}\)  &  0.004 &  4.02\(_{1.44}\) & 0.029\\
\midrule
FC1       & 9.56\(_{4.98}\)  & 0.010 & 9.72\(_{ 5.16}\) & 0.034 \\
FC2       & 5.68\(_{2.64}\)  & 0.021 & 9.72\(_{ 5.16}\) & 0.242 \\
\midrule
Total       & 9.53\(_{4.72}\)  & 0.004 & 4.66\(_{1.11}\) & 0.040 \\
\bottomrule
\end{tabular}
\caption{\textbf{OPT-125M Profiling Breakdown --} Disaggregating the profiling metrics for different layer types on OPT-125M. }
\label{tab:profiling-breakdown}
\end{table}

\begin{figure}[t!]
  \centering
   \includegraphics[width=\linewidth]{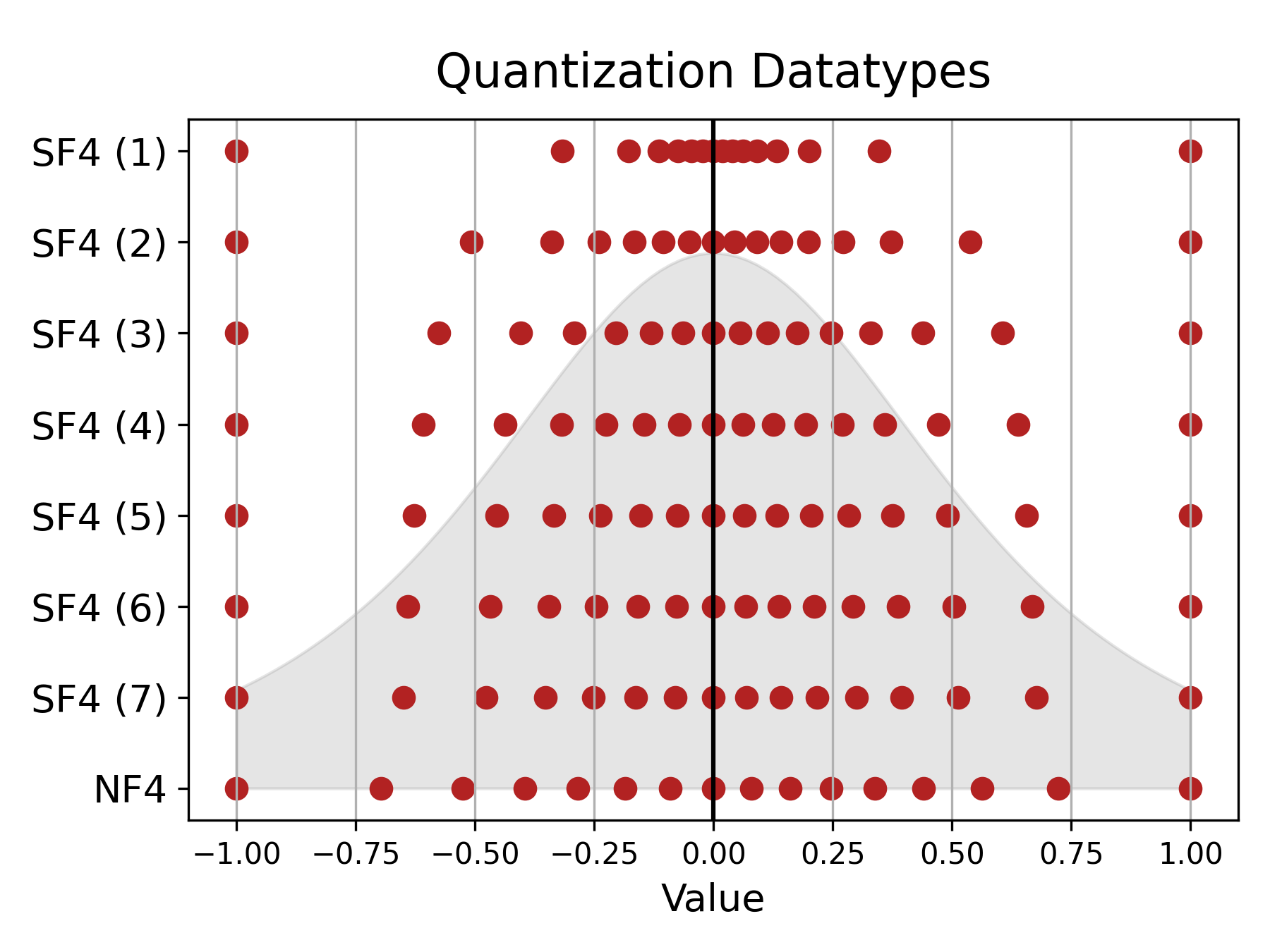}
   \vspace{-25pt}
   \caption{\textbf{Degrees of Freedom -- } Higher degrees of freedom lead to datatypes with more spread, and in the limit, SF4 approaches NF4. Most distributions have degrees of freedom close to 5, and therefore the SF4 ($\nu=5$) datatype is used throughout Section~\ref{sec:experiments}.}
   \label{fig:student-degrees}
\end{figure}

For weights and activation profiling, we use Huggingface transformers, PyTorch torchvision, and the timm package to load models. 
We chose the models holistically based on historical significance, current popularity, architectural types, and diversity across tasks.
This leads to including LLMs, BERT-like transformers, CNNs, RNNs, and diffusion models.

To profile the model, we iterate through the model modules and filter for nn.Linear, nn.Conv1D, and nn.Conv2D.
If the weight tensors are extremely large containing hundreds of millions of entries, we randomly downsample since small studies showed this did not significantly affect the profiling results.
For activation profiling, we use randomly generated inputs with the appropriate shape to match the current model.

Table~\ref{tab:profiling-additional} shows all the model profiling data, comparing between Student's t-distributions and normal distributions.
It lists the mean and variance for the degrees of freedom $\nu$ calculated across layers within the model.
In addition, it shows the difference between two Kolmogorov-Smirnov distances: the first is between the profiled distributions and the best-fitting normal distribution, and the second with respect to the best-fitting Student's t-distribution.
A positive difference between the normal and t-distribution distances indicates that the t-distribution is closer, and therefore it better represents the profiled data.

The degrees of freedom and KS-$\Delta$ are shown for both the weights and activations.
Overall, the activations typically have smaller degrees of freedom.
For example, BLOOM-7B has an average of 10.13 for its weights and 4.51 for its activations, and FLAN-T5 has 13.47 for its weights and 5.34 for its activations.
The degrees of freedom and KS-$\Delta$ are also very correlated, since a high degree of freedom indicates a distribution closer to normal.
Only the models with $\nu > 10$ have a negative KS-$\Delta$, which indicates this is a useful intuitive cutoff for classifying a distribution as normal.

In addition, we disaggregate the data across layer types, e.g. separating the attention layers from the linear layers in transformers.
This analysis is shown in Table~\ref{tab:profiling-breakdown} for the OPT-125M model, which separately averages the degrees of freedom and KS-$\Delta$ for different layer types.
It shows some differences between layer types, with FC2 having the lowest $\nu$, yet overall most layers are similar within their variance.

\section{Weight-Only}
\label{sec:weight-only}

\setlength{\tabcolsep}{3.5pt} 
\renewcommand{\arraystretch}{0.8} 
\begin{table*}[t!]
\centering
\footnotesize
\begin{tabular}{rrrrrrrrrrrrrrrr}
\toprule
 & & \multicolumn{2}{c}{\textbf{Mistral-7B}} & \multicolumn{2}{c}{\textbf{OPT-1B}} & \multicolumn{2}{c}{\textbf{OPT-6.7B}} & \multicolumn{2}{c}{\textbf{LLaMA2-7B}} & \multicolumn{2}{c}{\textbf{Phi-2}} & \multicolumn{2}{c}{\textbf{BLOOM-7B}} & \multicolumn{2}{c}{\textbf{Yi-6B}}\\
\midrule
Calib. Method &  & None & MSE & None & MSE & None & MSE & None & MSE & None & MSE & None & MSE & None & MSE\\
\midrule
\multirow{9}{*}{WikiText-2 ↓} & FP32  & 18.01 & 18.01 & 16.41 & 16.41 & 12.28 & 12.28 & 8.79 & 8.79 & 11.05 & 11.05 & 14.71 & 14.71 & 10.21 & 10.21\\
\cmidrule(lr){2-16}
& NF4   & 19.80  & 19.36 & 17.17 & 17.13 & 12.73 & 12.75 & \textbf{9.11} & 9.12 & 11.89 & 11.89 & \textbf{14.94} & \textbf{14.74} & 10.36 & 10.47\\
& SF4   &\textbf{19.09} & \textbf{19.34} & \textbf{ 17.11} & \textbf{17.10} & \textbf{12.67} & \textbf{12.66} & 9.16  &\textbf{ 9.10} & \textbf{11.83} & \textbf{11.84} & 14.96 & 14.84 & \textbf{10.34} & \textbf{10.36}\\
\cmidrule(lr){2-16}
& INT4  & 20.17 & 20.81 & 18.28 & 18.02 & 13.27 & 13.20 & 9.33  & 9.71 & 12.41 & 12.81 & 15.16 & 15.25 & 10.71 & 11.34\\
\cmidrule(lr){2-16}
& E2M1-I& 20.07 & 20.55 & 17.86  & 18.00 & 12.92 & 12.96 & 9.37 & 9.74 & 12.19 & 12.38 & 15.18 & 15.16 & 10.69 & 11.34\\
& E2M1-B& 20.93 & 21.17 & 18.34 & 18.15 & 13.11 & 13.19 & 9.43 & 9.89 & 12.37 & 12.64 & 15.22 & 15.26 & 10.76 & 11.54\\
& E2M1  & 19.76 & \textbf{19.27} & 17.24 & 17.25 & 12.78 & 12.79 & 9.17 & 9.21 & 11.97 & 11.99 & 15.01 & 15.18 & 10.42 & 10.54\\
& + SR  & 20.25 & 20.25 & 17.62 & 17.62 & 13.06 & 13.06 & 9.84 & 9.84 & 12.58 & 12.58 & 15.95 & 15.82 & 11.60 & 11.54\\
& + SP  & \textbf{19.38} & 19.47 & \textbf{17.19} & \textbf{17.18} & \textbf{12.76} & \textbf{12.77} & \textbf{9.13} &\textbf{ 9.20} & \textbf{11.92} & \textbf{11.96} & \textbf{14.98} & \textbf{14.89} & \textbf{10.37} & \textbf{10.29}\\
& E3M0  & 20.25 & 21.93 & 18.29 & 18.41 & 13.31 & 13.91 & 9.87 & 10.06 & 12.74 & 12.92 & 15.61 & 15.71 & 11.42 & 11.43\\
\cmidrule(lr){2-16}
& APoT4 & 19.13 & \textbf{19.23} & 17.47 & 17.42 & 12.84 & 12.88 & 9.15 & \textbf{9.27} & 12.09 & 12.17 & 15.02 &  14.98 & 10.46 & 10.49\\
& + SP & \textbf{18.93} & 19.32 & \textbf{17.40} & \textbf{17.32} & \textbf{12.80} & \textbf{12.85 }& \textbf{9.11} & 9.41 & \textbf{11.98} & \textbf{12.06} & \textbf{14.99} &  \textbf{14.92} & \textbf{10.40} & \textbf{10.39}\\
\bottomrule
\end{tabular}
\caption{\textbf{Weight-Only WikiText-2 --} All models evaluated with weight-only sub-channel quantization with block size 128. Student Float (SF4) typically outperforms NF4, and the super normal variants (SR and SP) often improve the model performance over E2M1.}
\label{tab:woq-wikitext}
\end{table*}

Table~\ref{tab:woq-wikitext} shows the additional evaluations across models on WikiText-2.
As a measure of perplexity, this is most sensitive metric to model changes, as others tend to mask their changes through a classification problem (e.g. multiple choice).
This table shows consistent improvement with SF4 over NF4 across models with the exception of BLOOM-7B.
Results are shown with and without MSE calibration.

\begin{table}[t]
\centering
\footnotesize
\small
\setlength{\tabcolsep}{3pt}
\begin{tabular}{@{}rrrrrrrr@{}}
\toprule
& \textbf{EN ↑} & \textbf{FR ↑} & \textbf{DE ↑} & \textbf{IT ↑} & \textbf{ES ↑} & \textbf{Wiki ↓} \\
\midrule
FP32      & 73.92 & 50.69 & 39.51 &  46.09 & 43.57 & 8.791 \\
\midrule
NF4      & \textbf{73.20} & 48.20 & 37.53 & 44.50 & 42.67 & \textbf{9.105} \\
SF4 & 72.35 & \textbf{48.79} & \textbf{38.54} & \textbf{44.81} & \textbf{44.44} & 9.163 \\
\midrule
INT4     & 72.06 & 47.45 & 37.26 & 42.87 & 42.60 & 9.333 \\
\midrule
E2M1-I & 71.43 & 47.43 & 37.07 & 42.48 & 42.05 & 9.366 \\
E2M1-B & 70.75 & 47.41 & 36.54 & 42.11 & 41.02 & 9.427 \\
E2M1 & 71.65 & \textbf{47.49} & 37.05 & \textbf{42.91} & \textbf{42.50} & 9.168 \\
+ SR & 71.07 & 45.27 & 35.14 & 41.45 & 39.36 & 9.842 \\
+ SP & \textbf{71.65} & 47.00 & \textbf{37.36} & 42.87 & 42.01 & \textbf{9.131} \\
E3M0       & 69.92 & 45.37 & 35.20          & 42.05 & 40.68  & 9.868 \\
\midrule
APoT4       & 72.77 & \textbf{48.98} & \textbf{37.88} & \textbf{45.16} & 41.53 & 9.149 \\
+ SP    & \textbf{73.22} & 48.75 & 37.55 & 44.34 & \textbf{41.57} & \textbf{9.109} \\

\bottomrule
\end{tabular}%
\vspace{-5pt}
\caption{\textbf{LLaMA2-7B Multi-Lingual --}  LLaMA2-7B comparison across multi-lingual LAMBADA tasks and WikiText-2. SF4 outperforms NF4 on lookup datatypes, and E2M1 with subnormal and super-precision outperforms other FP4 datatypes.}
\label{tab:language-multi}
\end{table}

Table~\ref{tab:language-multi} shows the results of LLaMA2-7B on a multi-lingual version of the LAMBADA dataset.
It reinforces the previous trends, which SF4 typically achieving higher accuracy and E2M1 with and without super-precision outperform other datatypes.

\section{Student Float}
\label{sec:sf4-appendix}

\begin{figure}[t!]
  \centering
   \includegraphics[width=.90\linewidth]{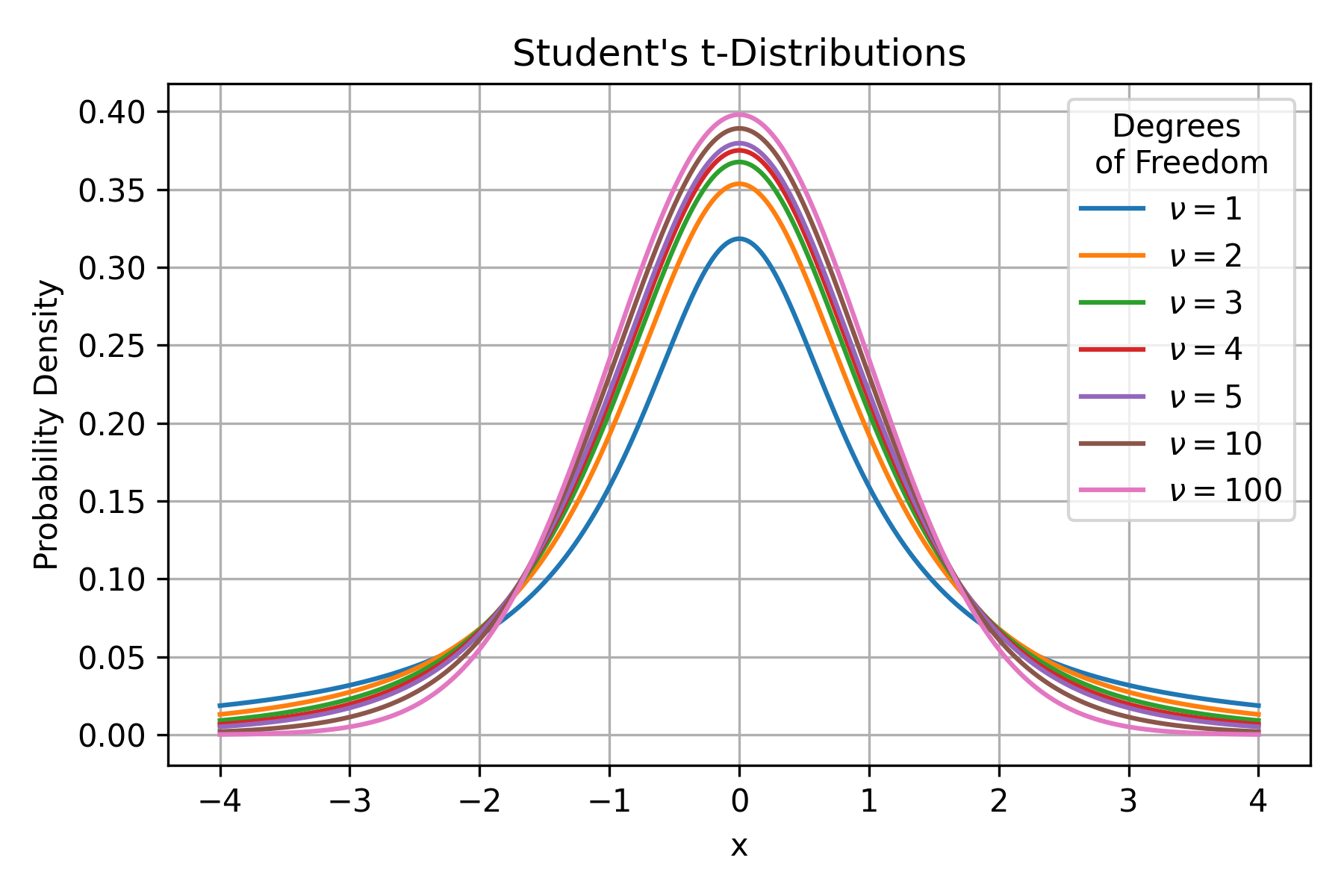}
   \vspace{-15pt}
   \caption{\textbf{t-Distributions -- } Increasing the degrees of freedom, $\nu$, leads to more probability mass in the center, and less at the edges of the distribution. This leads to more representation in the center of the SF4 datatype, and in the limit, the NF4 datatype.}
   \label{fig:t-distributions}
\end{figure}

Figure~\ref{fig:student-degrees} shows that SF4 converges to NF4 as its degrees of freedom increase to infinity.
This allows testing for gradually denser datatypes toward NF4 and making comparisons to the corresponding degrees of freedom in the profiling results in Table~\ref{tab:profiling-additional}.
Overall, on average models approximately have $\nu = 5$, which leads to the highest accuracy results across tasks.

In addition, Figure~\ref{fig:t-distributions} shows the direct effect of increasing the degrees of freedom ($\nu$) on the curvature of the Student's t-distribution. Higher $\nu$ leads to wider peaks and thinner tails.

\section{Datatype Values}
\label{sec:datatypes}

\begin{figure}[t]
  \centering
   \includegraphics[width=\linewidth]{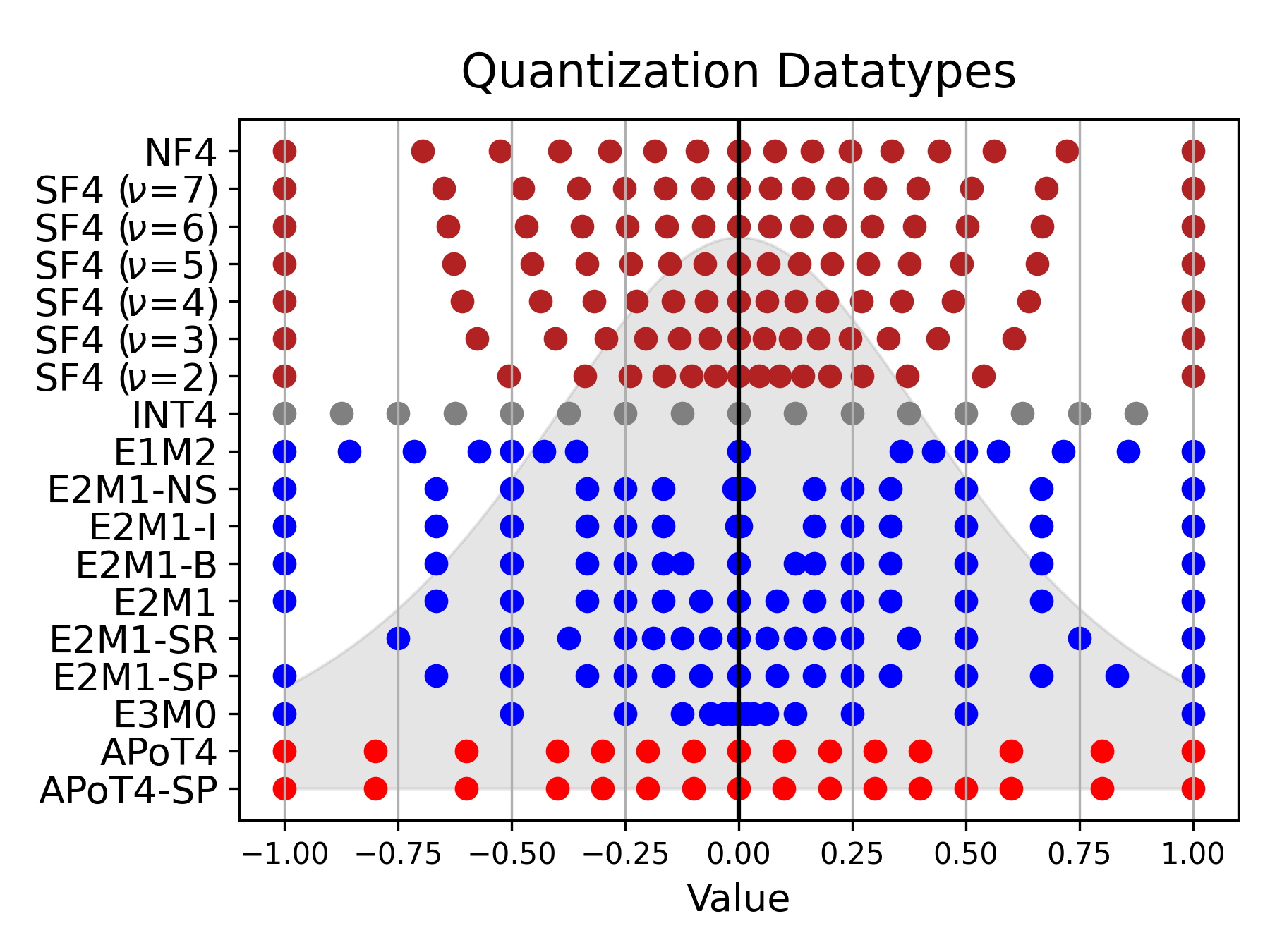}
   \vspace{-25pt}
   \caption{\textbf{Datatype Shapes --} The shapes of all considered datatypes, including lookup datatypes, integer, floating-point, and APoT~\cite{li2020apot}.}
   \vspace{-10pt}
   \label{fig:datatypes-all}
\end{figure}
\begin{table*}[t!]
\centering
\footnotesize
\setlength{\tabcolsep}{2pt}
\begin{tabular}{rrrrrrrrrrrrrrrrr}
\toprule
\textbf{Datatype} & \multicolumn{16}{c}{\textbf{Values}} \\
\midrule
NF4 & -1.000 & -0.696 & -0.525 & -0.395 & -0.284 & -0.185 & -0.091 & 0.000 & 0.080 & 0.161 & 0.246 & 0.338 & 0.441 & 0.563 & 0.723 & 1.000 \\
SF4 ($\nu=3$) & -1.000 & -0.576 & -0.404 & -0.292 & -0.205 & -0.131 & -0.064 & 0.000 & 0.056 & 0.114 & 0.176 & 0.246 & 0.330 & 0.439 & 0.606 & 1.000 \\
SF4 ($\nu=4$) & -1.000 & -0.609 & -0.436 & -0.318 & -0.225 & -0.145 & -0.071 & 0.000 & 0.062 & 0.126 & 0.194 & 0.270 & 0.359 & 0.472 & 0.638 & 1.000 \\
SF4 ($\nu=5$) & -1.000 & -0.628 & -0.455 & -0.334 & -0.237 & -0.153 & -0.075 & 0.000 & 0.066 & 0.133 & 0.205 & 0.284 & 0.376 & 0.491 & 0.657 & 1.000 \\
SF4 ($\nu=6$) & -1.000 & -0.640 & -0.467 & -0.345 & -0.246 & -0.158 & -0.078 & 0.000 & 0.068 & 0.138 & 0.212 & 0.293 & 0.387 & 0.504 & 0.669 & 1.000 \\
\midrule
INT4 & -8.000 & -7.000 & -6.000 & -5.000 & -4.000 & -3.000 & -2.000 & -1.000 & 0.000 & 1.000 & 2.000 & 3.000 & 4.000 & 5.000 & 6.000 & 7.000 \\
\midrule
E2M1-I & -6.000 & -4.000 & -3.000 & -2.000 & -1.500 & -1.000 & -0.062 & 0.000 & 0.062 & 1.000 & 1.500 & 2.000 & 3.000 & 4.000 & 6.000 & \\
E2M1-B & -12.000 & -8.000 & -6.000 & -4.000 & -3.000 & -2.000 & -0.062 & 0.000 & 0.062 & 2.000 & 3.000 & 4.000 & 6.000 & 8.000 & 12.000 & \\
E2M1-NS & -6.000 & -4.000 & -3.000 & -2.000 & -1.500 & -1.000 & -0.750 & 0.000 & 0.750 & 1.000 & 1.500 & 2.000 & 3.000 & 4.000 & 6.000 & \\
E2M1 & -6.000 & -4.000 & -3.000 & -2.000 & -1.500 & -1.000 & -0.500 & 0.000 & 0.500 & 1.000 & 1.500 & 2.000 & 3.000 & 4.000 & 6.000 & \\
+ SR & -6.000 & -4.000 & -3.000 & -2.000 & -1.500 & -1.000 & -0.500 & 0.000 & 0.500 & 1.000 & 1.500 & 2.000 & 3.000 & 4.000 & 6.000 & 8.000 \\
+ SP & -6.000 & -4.000 & -3.000 & -2.000 & -1.500 & -1.000 & -0.500 & 0.000 & 0.500 & 1.000 & 1.500 & 2.000 & 3.000 & 4.000 & 5.000 & 6.000 \\
E3M0 & -16.000 & -8.000 & -4.000 & -2.000 & -1.000 & -0.500 & -0.250 & 0.000 & 0.250 & 0.500 & 1.000 & 2.000 & 4.000 & 8.000 & 16.000 & \\
\midrule
APoT4 & -1.000 & -0.800 & -0.600 & -0.400 & -0.300 & -0.200 & -0.100 & 0.000 & 0.100 & 0.200 & 0.300 & 0.400 & 0.600 & 0.800 & 1.000 & \\
+ SP & -1.000 & -0.800 & -0.600 & -0.400 & -0.300 & -0.200 & -0.100 & 0.000 & 0.100 & 0.200 & 0.300 & 0.400 & 0.500 & 0.600 & 0.800 & 1.000 \\
\bottomrule
\end{tabular}
\caption{\textbf{Quantized Datatype Values --} The specific values for each datatype across lookup, integer, floating-point, and alternative formats. Some datatypes have only 15 values, as opposed to 16 ($2^4$), since they include a dedicated sign bit, which leads to representations for positive and negative zero. The Student Float (SF4) formats include versions for different degrees of freedom ($\nu$), which cluster values in different ways. For floating-point formats, the Intel~\cite{shen2023efficient} (I-E2M1) and bitsandbytes~\cite{dettmers2022llmint8} (B-E2M1) versions are included as references too. Additive Powers of Two (APoT)~\cite{li2020apot} is also shown which performs the sum of two logarithmic numbers. Finally, the super-precision (SP), super-range (SR), and no subnormal (NS) variants are shown for some of these formats.}
\label{table:quantized_values}
\end{table*}

This section lists the values for all the datatypes used in the evaluations in Section~\ref{sec:experiments} and Section~\ref{sec:performance}.
In addition, it shows all of the datatypes in the same figure, including the lookup datatypes, integer, floating-point, and APoT variants.

\section{Additive Powers-of-Two}
\label{sec:apot}

\begin{figure}[t]
  \centering
   \includegraphics[width=.90\linewidth]{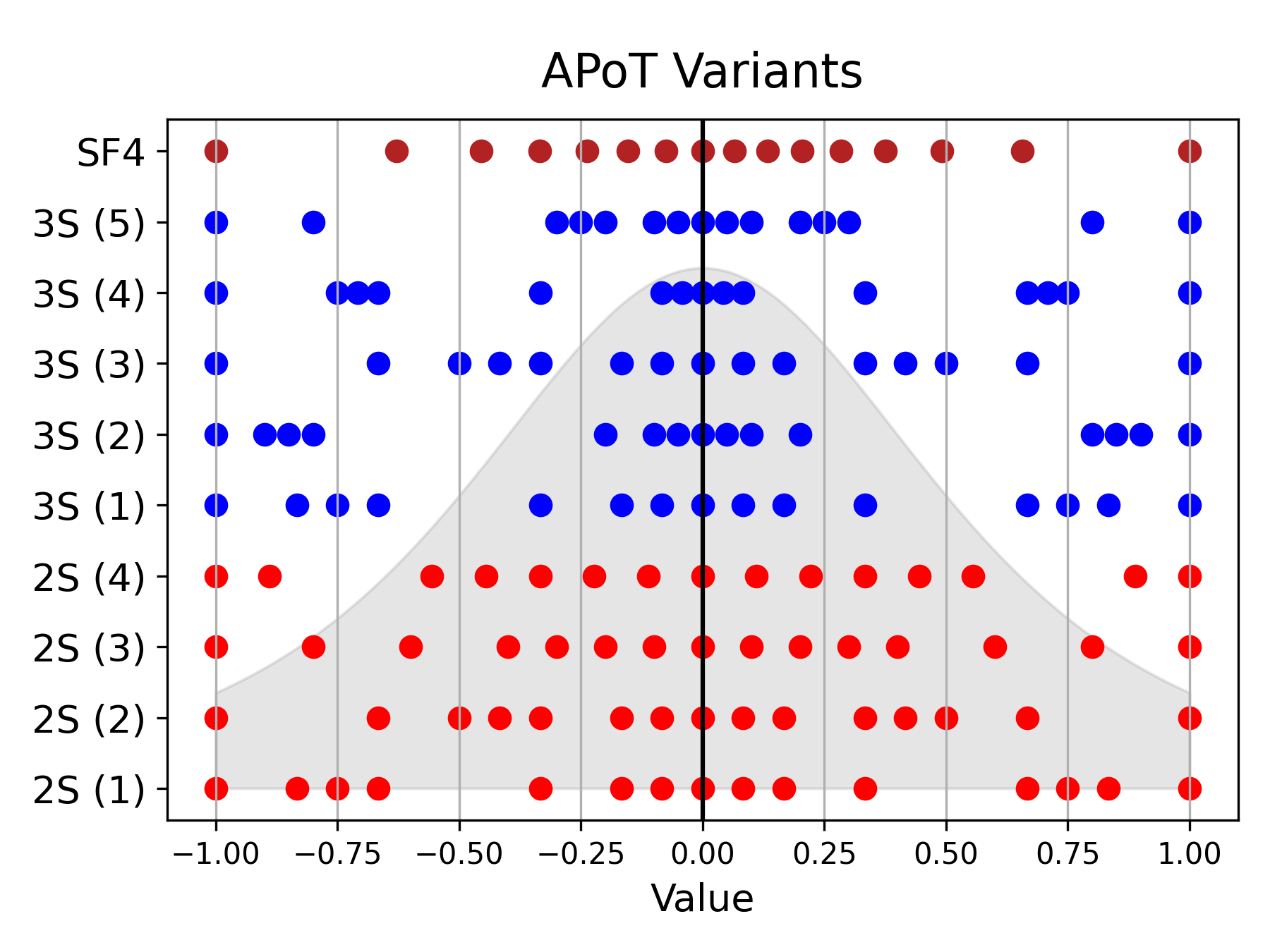}
   \vspace{-10pt}
   \caption{\textbf{APoT4 Variants --} Comparison across APoT4 variants with two sets (2S) and three sets (3S), where each datatype is constructed by all possible sums by taking one value from each set. For example, the 2S (3) variant used in Section~\ref{sec:experiments}, uses the sets $S_1 \in \{0, 2^{-1}, 2^{-2}, 2^{-4}\}$ and $ S_2 \in \{0, 2^{-3}\}$. The values to construct the sets are always drawn from $\{0, 2^{-1}, 2^{-2}, 2^{-4}\}$. SF4 is shown for reference.}
   \vspace{-10pt}
   \label{fig:apot-values}
\end{figure}

The Additive Powers-of-Two method leads to a large search space of datatypes, where all the most reasonable variants are shown in Figure~\ref{fig:apot-values}.
These have been filtered to remove datatypes that lead to duplicate values (under-utilizing the bitspace) and different configurations that lead to the exact same datatype.
This figure shows that the 2S (3) variant best approximates the SF4 datatype, and therefore in this work we focus only on this variant.

\section{Additional Paretos}
\label{sec:additional-paretos}

\begin{figure*}[t]
  \centering
   \includegraphics[width=.95\linewidth]{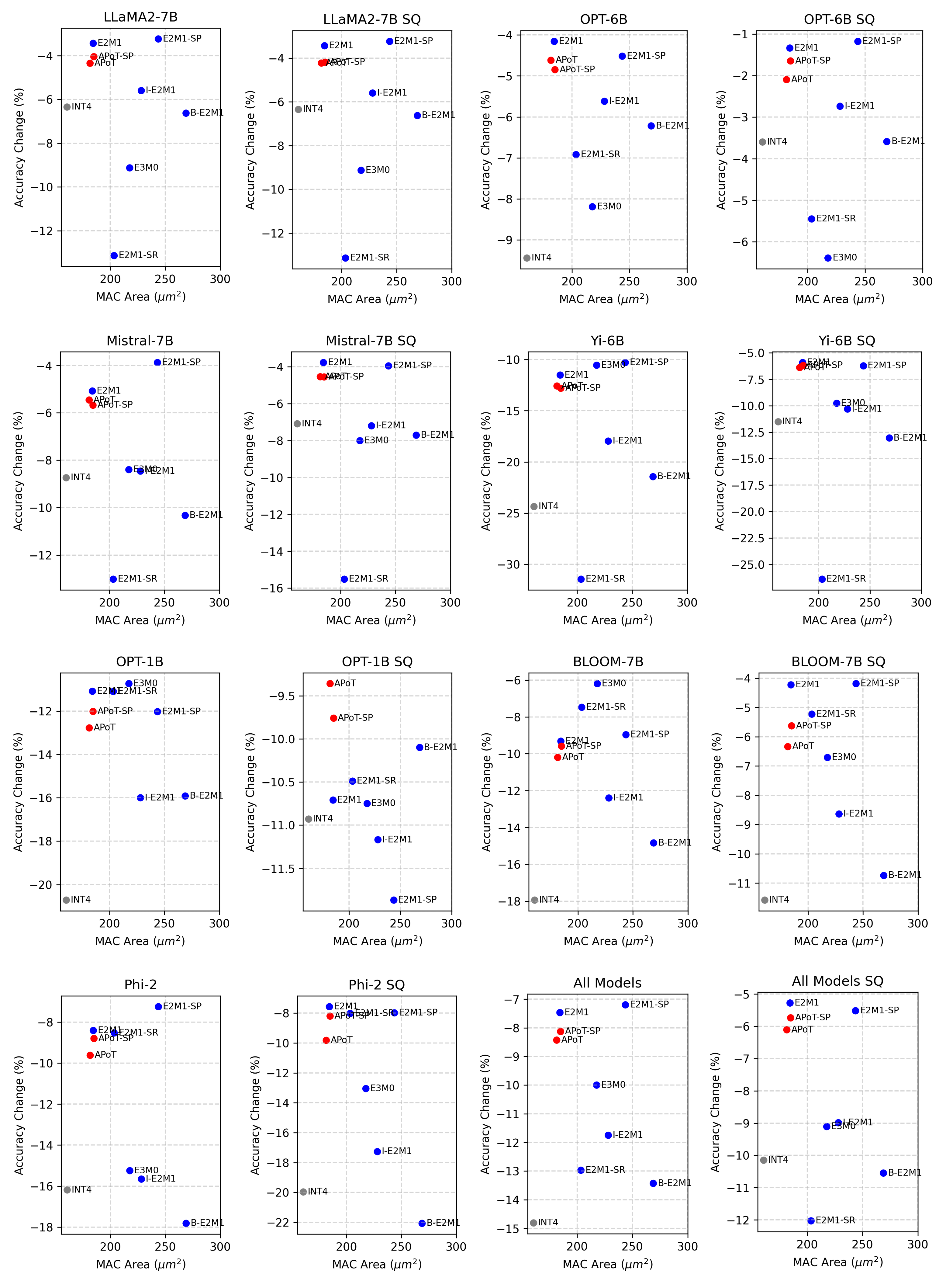}
   \vspace{-10pt}
   \caption{\textbf{All Model Paretos -- } Relative accuracy change from unquantized baselines averaged across LAMBADA, HellaSwag, Winogrande, PIQA, BoolQ, and ARC-c. All models are quantized with W4A4 subchannel quantization with SmoothQuant~\cite{xiao2023smoothquant} included on models with the SQ label. }
   \vspace{-10pt}
   \label{fig:paretos}
\end{figure*}

This section includes all of the Pareto-curves for Mistral-7B, OPT-1B, OPT-6.7B, LLaMA2-7B, Phi-2, BLOOM-7B, and Yi-6B evaluated across LAMBADA, HellaSwag, Winogrande, PIQA, BoolQ, and ARC-c. 
The y-axis represents the average relative accuracy change from floating-point, and the x-axis is the corresponding MAC area for the datatype.

\clearpage
\section{Additional Tables}
\label{sec:additional-tables}

\begin{table}[th]
\centering
\footnotesize
\small
\setlength{\tabcolsep}{2pt}
\begin{tabular}{@{}rrrrrrr@{}} 
\toprule
Metric & \textbf{LAMB} & \textbf{Hella} & \textbf{Wino} & \textbf{PIQA} & \textbf{BoolQ} & \textbf{ARC-c} \\ 
\midrule
BF16      & 73.92 & 57.14 & 69.14 & 78.07 & 77.74 & 43.43 \\
\midrule
NF4       & 72.35 & 56.55 & 69.53 & 76.99 & 77.40 & 42.49 \\
SF4 & 73.20 & 56.81 & 69.06 & 77.69 & 78.56 & 43.34 \\
\midrule
INT4      & 72.06 & 56.53 & 69.14 & 77.31 & 76.76 & 42.92 \\
\midrule
I-E2M1    & 71.43 & 56.50 & 68.90 & 77.80 & 77.06 & 42.66 \\
B-E2M1    & 70.75 & 56.54 & 68.98 & 77.58 & 76.73 & 43.34 \\
E2M1      & 71.65 & 56.69 & 69.53 & 77.97 & 78.13 & 42.49 \\
\texttt{+} SR      & 71.07 & 54.66 & 66.85 & 76.77 & 73.55 & 42.41 \\
\texttt{+} SP      & 71.65 & 56.84 & 69.43 & 77.99 & 78.26 & 42.49 \\
E3M0      & 69.92 & 54.61 & 67.64 & 76.55 & 75.32 & 39.59 \\
\midrule
APoT4      & 72.77 & 56.27 & 68.27 & 78.07 & 77.55 & 43.17 \\
\texttt{+} SP      & 73.22 & 56.56 & 68.59 & 77.69 & 77.68 & 43.86 \\
\bottomrule
\end{tabular}%
\vspace{-5pt}
\caption{\textbf{LLaMA-7B Weight-Only Subchannel 128}}
\label{tab:llama2_7b_evaluation}
\end{table}

\begin{table}[th]
\centering
\footnotesize
\small
\setlength{\tabcolsep}{2pt}
\begin{tabular}{@{}rrrrrrr@{}} 
\toprule
& \textbf{LAMB} & \textbf{Hella} & \textbf{Wino} & \textbf{PIQA} & \textbf{BoolQ} & \textbf{ARC-c} \\ 
\midrule
FP32      & 62.57 & 55.84 & 75.45 & 78.78 & 83.21 & 52.56 \\
\midrule
NF4       & 60.47 & 54.66 & 75.22 & 77.42 & 82.81 & 50.85 \\
SF4 & 61.28 & 54.75 & 75.30 & 78.13 & 80.76 & 52.56 \\
\midrule
INT4      & 58.59 & 54.51 & 75.61 & 77.69 & 79.14 & 51.02 \\
\midrule
I-E2M1    & 58.20 & 54.06 & 74.59 & 77.69 & 82.45 & 51.28 \\
B-E2M1    & 58.32 & 54.07 & 75.22 & 77.04 & 82.32 & 50.85 \\
E2M1      & 59.95 & 54.83 & 76.24 & 77.09 & 83.06 & 51.96 \\
+ SR      & 63.24 & 53.32 & 75.06 & 78.40 & 81.38 & 50.17 \\
+ SP      & 61.73 & 55.06 & 76.01 & 76.99 & 83.21 & 52.73 \\
E3M0      & 54.96 & 52.18 & 74.59 & 78.56 & 80.86 & 50.43 \\
\midrule
APoT4      & 59.62 & 54.50 & 74.35 & 77.91 & 81.35 & 52.82 \\
+ SP      & 61.09 & 54.66 & 74.27 & 78.35 & 81.71 & 52.90 \\
\bottomrule
\end{tabular}%
\vspace{-5pt}
\caption{\textbf{Phi-2 Weight-Only Subchannel 128}}
\label{tab:phi2_evaluation}
\end{table}

\begin{table}[th]
\centering
\footnotesize
\small
\setlength{\tabcolsep}{2pt}
\begin{tabular}{@{}rrrrrrr@{}} 
\toprule
Metric & \textbf{LAMB} & \textbf{Hella} & \textbf{Wino} & \textbf{PIQA} & \textbf{BoolQ} & \textbf{ARC-c} \\
\midrule
FP32      & 75.92 & 61.22 & 73.88 & 80.58 & 83.58 & 50.43 \\
\midrule
NF4       & 74.97 & 60.90 & 72.93 & 80.30 & 82.84 & 49.74 \\
SF4 & 75.90 & 60.73 & 73.80 & 80.63 & 83.09 & 49.40 \\
\midrule
INT4      & 73.92 & 60.59 & 73.80 & 80.36 & 82.23 & 49.32 \\
\midrule
I-E2M1    & 74.17 & 60.41 & 72.45 & 80.36 & 82.84 & 48.98 \\
B-E2M1    & 73.98 & 60.36 & 72.22 & 80.09 & 82.48 & 48.81 \\
E2M1      & 74.75 & 60.57 & 73.16 & 80.14 & 82.29 & 48.55 \\
\texttt{+} SR      & 72.95 & 59.07 & 73.56 & 79.65 & 82.84 & 47.95 \\
\texttt{+} SP      & 75.41 & 60.96 & 72.93 & 80.36 & 83.46 & 47.78 \\
E3M0      & 74.23 & 58.76 & 72.22 & 79.71 & 81.99 & 46.42 \\
\midrule
APoT4      & 75.41 & 60.89 & 73.95 & 80.30 & 83.09 & 47.44 \\
\texttt{+} SP      & 75.12 & 61.05 & 73.09 & 80.20 & 83.03 & 48.21 \\
\bottomrule
\end{tabular}
\vspace{-5pt}
\caption{\textbf{Mistral-7B Weight-Only Subchannel 128}}
\label{tab:mistral_7b_evaluation}
\end{table}

\begin{table}[th]
\centering
\footnotesize
\small
\setlength{\tabcolsep}{2pt}
\begin{tabular}{@{}rrrrrrr@{}} 
\toprule
Metric & \textbf{LAMB} & \textbf{Hella} & \textbf{Wino} & \textbf{PIQA} & \textbf{BoolQ} & \textbf{ARC-c} \\
\midrule
FP32      & 68.27 & 55.40 & 70.96 & 77.64 & 75.50 & 46.25 \\
\midrule
NF4       & 67.46 & 54.81 & 71.03 & 77.26 & 78.47 & 44.97 \\
SF4 & 67.84 & 54.75 & 70.80 & 77.15 & 76.97 & 45.14 \\
\midrule
INT4      & 64.93 & 54.51 & 68.75 & 77.31 & 75.41 & 44.37 \\
\midrule
I-E2M1    & 64.39 & 54.48 & 71.11 & 77.26 & 75.81 & 44.71 \\
B-E2M1    & 63.92 & 54.56 & 70.56 & 77.09 & 75.32 & 44.20 \\
E2M1      & 66.74 & 54.52 & 69.85 & 76.71 & 76.57 & 45.05 \\
\texttt{+} SR      & 59.97 & 52.95 & 67.80 & 75.90 & 76.18 & 43.52 \\
\texttt{+} SP      & 67.38 & 54.83 & 70.56 & 76.71 & 76.27 & 46.50 \\
E3M0      & 65.15 & 52.48 & 68.90 & 76.33 & 73.82 & 41.81 \\
\midrule
APoT4      & 68.21 & 55.08 & 70.24 & 77.69 & 77.49 & 45.73 \\
\texttt{+} SP      & 68.14 & 55.25 & 70.88 & 77.58 & 77.34 & 45.39 \\
\bottomrule
\end{tabular}
\vspace{-5pt}
\caption{\textbf{Yi-6B Weight-Only Subchannel 128}}
\label{tab:yi_6b_evaluation}
\end{table}

\begin{table}[th]
\centering
\footnotesize
\small
\setlength{\tabcolsep}{2pt}
\begin{tabular}{@{}rrrrrrr@{}} 
\toprule
Metric & \textbf{LAMB} & \textbf{Hella} & \textbf{Wino} & \textbf{PIQA} & \textbf{BoolQ} & \textbf{ARC-c} \\ 
\midrule
FP32      & 57.64 & 46.49 & 64.56 & 72.69 & 62.81 & 30.29 \\
\midrule
NF4       & 57.03 & 45.47 & 62.98 & 72.96 & 63.46 & 30.38 \\
SF4 & 57.77 & 45.43 & 64.25 & 72.25 & 62.87 & 29.86 \\
\midrule
INT4      & 56.08 & 45.31 & 63.54 & 73.12 & 63.55 & 29.44 \\
\midrule
I-E2M1    & 55.75 & 45.66 & 63.38 & 72.80 & 63.24 & 29.95 \\
B-E2M1    & 55.64 & 45.47 & 62.90 & 72.96 & 63.21 & 30.20 \\
E2M1      & 56.51 & 45.26 & 63.30 & 72.63 & 63.43 & 30.12 \\
\texttt{+} SR      & 50.18 & 44.56 & 62.75 & 72.63 & 61.44 & 30.63 \\
\texttt{+} SP      & 56.86 & 45.41 & 63.46 & 72.74 & 63.46 & 30.03 \\
E3M0      & 56.47 & 44.36 & 61.25 & 72.47 & 63.67 & 29.78 \\
\midrule
APoT4      & 57.02 & 45.30 & 63.85 & 72.96 & 62.57 & 29.86 \\
\texttt{+} SP      & 57.13 & 45.46 & 63.22 & 72.47 & 62.72 & 29.86 \\
\bottomrule
\end{tabular}%
\vspace{-5pt}
\caption{\textbf{BLOOM-7B Weight-Only Subchannel 128}}
\label{tab:bloom_7b_evaluation}
\end{table}

\begin{table}[th]
\centering
\footnotesize
\small
\setlength{\tabcolsep}{2pt}
\begin{tabular}{@{}rrrrrrr@{}} 
\toprule
Metric & \textbf{LAMB} & \textbf{Hella} & \textbf{Wino} & \textbf{PIQA} & \textbf{BoolQ} & \textbf{ARC-c} \\
\midrule
FP32      & 67.69 & 50.49 & 65.43 & 76.28 & 66.06 & 30.72 \\
\midrule
NF4       & 67.88 & 49.34 & 64.25 & 76.22 & 65.99 & 30.63 \\
SF4 & 68.02 & 49.58 & 64.96 & 75.90 & 64.04 & 30.03 \\
\midrule
INT4      & 63.92 & 49.02 & 63.93 & 75.63 & 65.23 & 31.23 \\
\midrule
I-E2M1    & 67.49 & 49.44 & 64.17 & 76.22 & 65.84 & 30.20 \\
B-E2M1    & 66.97 & 49.42 & 63.06 & 76.55 & 67.06 & 31.14 \\
E2M1      & 67.84 & 49.15 & 64.17 & 76.06 & 66.02 & 30.63 \\
\texttt{+} SR      & 67.26 & 48.48 & 64.48 & 75.14 & 63.46 & 29.44 \\
\texttt{+} SP      & 67.24 & 49.29 & 63.77 & 76.17 & 65.96 & 30.38 \\
E3M0      & 62.64 & 48.16 & 63.38 & 74.65 & 65.96 & 30.12 \\
\midrule
APoT4      & 66.08 & 49.64 & 64.64 & 75.79 & 65.02 & 30.63 \\
\texttt{+} SP      & 65.92 & 49.59 & 64.96 & 75.95 & 64.31 & 31.06 \\
\bottomrule
\end{tabular}
\vspace{-5pt}
\caption{\textbf{OPT-6B Weight-Only Subchannel 128}}
\label{tab:opt_6b_evaluation}
\end{table}

\begin{table}[t]
\centering
\footnotesize
\small
\setlength{\tabcolsep}{1.5pt}
\begin{tabular}{@{}crrrrrrrr@{}} 
\toprule
& & \textbf{LAMB} &  \textbf{Hella} &  \textbf{Wino} &  \textbf{PIQA} &  \textbf{BoolQ} & \textbf{ARC-c} \\
\midrule
\multirow{12}{*}{\raisebox{-1.5cm}{\hspace{-1em}\rotatebox[origin=c]{90}{No SmoothQuant}}}
& FP32 & 68.27 & 55.4 & 70.96 & 77.64 & 75.5 & 46.25  \\
\midrule
& NF4 & 51.17 & 51.34 & 63.77 & 74.21 & 71.93 & \textbf{40.70}  \\
& SF4 & \textbf{55.29} & \textbf{51.58} & \textbf{64.33} & \textbf{74.59} & \textbf{73.03} & 40.44  \\
\midrule
& INT4 & 31.4 & 46.14 & 56.2 & 71.49 & 58.84 & 34.81  \\
\midrule
& I-E2M1 & 42.36 & 48.89 & 60.14 & 71.93 & 64.16 & 36.77  \\
& B-E2M1 & 34.52 & 47.16 & 55.64 & 70.78 & 63.64 & 37.80  \\
& E2M1 & 49.62 & \textbf{50.93} & 63.61 & 73.23 & \textbf{72.02} & 40.19  \\
&  \texttt{+} SR & 23.50 & 41.69 & 55.33 & 65.13 & 63.12 & 25.94  \\
&  \texttt{+} SP & 48.13 & 50.80 & 63.77 & \textbf{74.21} & 66.61 & \textbf{40.36}  \\
& E3M0 & \textbf{59.07} & 49.19 & \textbf{64.80} & 73.07 & 69.97 & 38.48  \\
\midrule
& APoT4 & 47.18 & 50.42 & 62.35 & \textbf{74.48} & \textbf{69.05} & \textbf{41.21}  \\
&+ SP & \textbf{48.13} & \textbf{50.80} & \textbf{63.77} & 74.21 & 66.61 & 40.36 \\
\midrule
\midrule
\multirow{12}{*}{\raisebox{-1.5cm}{\hspace{-1em}\rotatebox[origin=c]{90}{SmoothQuant}}} 
& NF4 & 61.81 & 53.40 & 65.59 & 74.92 & 72.75 & 43.94  \\
& SF4 & \textbf{64.72} & \textbf{53.48} & \textbf{66.93} & \textbf{76.61} & \textbf{73.24} & \textbf{44.45}  \\
\midrule
& INT4 & 51.85 & 51.13 & 63.93 & 74.65 & 68.29 & 39.76  \\
\midrule
& I-E2M1 & 53.58 & 51.55 & 63.38 & 74.48 & 68.20 & 42.06  \\
& B-E2M1 & 51.39 & 50.93 & 62.27 & 73.78 & 67.25 & 38.23  \\
& E2M1 & \textbf{61.91} & 53.13 & 65.59 & \textbf{75.84} & 69.45 & \textbf{44.28}  \\
&  \texttt{+} SR & 34.97 & 44.82 & 57.46 & 65.51 & 65.47 & 26.62  \\
&  \texttt{+} SP & 59.25 & \textbf{53.37} & \textbf{66.69} & 75.35 & 70.70 & 43.94  \\
& E3M0 & 59.77 & 49.82 & 65.35 & 74.16 & \textbf{72.08} & 37.37  \\
\midrule
& APoT4 & 58.80 & 53.07 & \textbf{67.64} & 74.43 & \textbf{72.81} & 42.58  \\
&\texttt{+} SP & \textbf{59.25} & \textbf{53.37} & 66.69 & \textbf{75.35} & 70.70 & \textbf{43.94} \\
\bottomrule
\end{tabular}%
\vspace{-5pt}
\caption{ \textbf{Yi-6B W4A4 Subchannel 128}}
\label{tab:yi_qa}
\vspace{-10pt}
\end{table}
\begin{table}[t]
\centering
\footnotesize
\small
\setlength{\tabcolsep}{1.5pt}
\begin{tabular}{@{}crrrrrrrr@{}} 
\toprule
& & \textbf{LAMB} & \textbf{Hella} & \textbf{Wino} & \textbf{PIQA} & \textbf{BoolQ} & \textbf{ARC-c} \\
\midrule
\multirow{13}{*}{\raisebox{-1.5cm}{\hspace{-1em}\rotatebox[origin=c]{90}{No SmoothQuant}}}
    & FP32 & 57.64 & 46.49 & 64.56 & 72.69 & 62.81 & 30.29  \\
\midrule    
& NF4 & 44.23 & 42.69 & \textbf{59.12} & 69.86 & \textbf{60.55} & \textbf{29.18}  \\
& SF4 & \textbf{48.98} & \textbf{43.24} & 59.04 & \textbf{70.29} & 58.87 & 29.01  \\
\midrule
& INT4 & 31.15 & 39.91 & 54.38 & 67.79 & 54.16 & 26.88  \\
\midrule
& I-E2M1 & 41.8 & 42.04 & 55.33 & 68.72 & 57.22 & 27.65  \\
& B-E2M1 & 36.48 & 40.83 & 54.78 & 67.95 & 57.77 & 27.13  \\
& E2M1 & 44.21 & 42.37 & \textbf{59.51} & 70.02 & 59.51 & 28.16  \\
& + SR & 48.22 & 41.51 & 57.22 & 70.62 & 61.96 & \textbf{29.61}  \\
& + SP & 44.58 & \textbf{42.82} & 58.48 & \textbf{70.73} & 59.69 & 28.41  \\
& E3M0 & \textbf{52.55} & 42.48 & 56.51 & 70.24 & \textbf{62.48} & 29.27  \\
\midrule
& APoT4 & 40.15 & 41.95 & 58.88 & 70.40 & \textbf{60.98} & 28.41  \\
& +SP & \textbf{41.35} & \textbf{41.98} & 59.19 & 70.62 & 59.82 & \textbf{29.18}  \\
\midrule
\midrule
\multirow{13}{*}{\raisebox{-1.5cm}{\hspace{-1em}\rotatebox[origin=c]{90}{SmoothQuant}}}
& NF4 & 52.90 & 44.50 & 60.69 & 71.38 & 61.65 & 28.84  \\
& SF4 & \textbf{55.29} & \textbf{45.06} & \textbf{61.09} & \textbf{72.31} & \textbf{63.64} & \textbf{29.86}  \\
\midrule
& INT4 & 41.72 & 41.72 & 56.83 & 69.53 & 57.13 & 28.41  \\
\midrule
& I-E2M1 & 47.08 & 42.21 & 57.06 & 69.91 & 61.50 & 28.24  \\
& B-E2M1 & 43.76 & 41.06 & 56.67 & 69.86 & 61.13 & 27.30  \\
& E2M1 & \textbf{53.77} & \textbf{44.52} & \textbf{60.46} & \textbf{71.76} & 61.74 & 28.75  \\
& + SR & 52.94 & 42.11 & 58.41 & 71.06 & \textbf{63.30} & 29.44  \\
& + SP & 51.09 & 43.92 & 58.98 & 70.78 & 59.62 & \textbf{30.12}  \\
& E3M0 & 51.93 & 42.4 & 57.93 & 69.8 & 62.84 & 28.07  \\
\midrule
& APoT & 50.11 & 43.81 & 58.33 & 70.62 & 59.48 & 29.86  \\
    &+ SP & \textbf{51.09} & \textbf{43.92} & \textbf{58.98} & \textbf{70.78} & \textbf{59.62} & \textbf{30.12} \\
\bottomrule
\end{tabular}%
\vspace{-5pt}
\caption{ \textbf{BLOOM-7B W4A4 Subchannel 128}}
\label{tab:bloom_qa}
\vspace{-10pt}
\end{table}
\begin{table}[t]
\centering
\footnotesize
\small
\setlength{\tabcolsep}{1.5pt}
\begin{tabular}{@{}crrrrrrrr@{}} 
\toprule
& & \textbf{LAMB} & \textbf{Hella} & \textbf{Wino} & \textbf{PIQA} & \textbf{BoolQ} & \textbf{ARC-c} \\
\midrule
\multirow{12}{*}{\raisebox{-1.5cm}{\hspace{-1em}\rotatebox[origin=c]{90}{No SmoothQuant}}}
& FP32 & 73.92 & 57.14 & 69.14 & 78.07 & 77.74 & 43.43  \\
\midrule
& NF4 & \textbf{73.03} & \textbf{55.57} & \textbf{67.09} & 76.55 & \textbf{75.96} & 41.38  \\
& SF4 & 72.21 & 55.28 & 66.69 & \textbf{76.93} & 75.72 & \textbf{41.81}  \\
\midrule
& INT4 & 69.92 & 53.76 & 65.27 & 75.79 & 69.88 & 40.10  \\
\midrule
& I-E2M1 & 69.55 & 54.33 & 65.11 & 75.57 & 70.34 & 40.27  \\
& B-E2M1 & 68.31 & 53.65 & 62.43 & 74.81 & 70.0 & 40.27  \\
& E2M1 & 72.21 & \textbf{55.61} & \textbf{67.01} & \textbf{76.39} & \textbf{76.24} & \textbf{41.72}  \\
& + SR & 63.96 & 48.91 & 61.01 & 73.18 & 70.18 & 35.58  \\
& + SP & \textbf{72.64} & 54.79 & 66.61 & 76.66 & 73.88 & 41.38  \\
& E3M0 & 65.03 & 51.29 & 62.35 & 74.43 & 69.42 & 36.26  \\
\midrule
& APoT4 & \textbf{72.79} & \textbf{55.01} & 65.82 & 76.39 & \textbf{74.07} & 41.04  \\
&\texttt{+} SP & 72.64 & 54.79 & \textbf{66.61} & \textbf{76.66} & 73.88 & \textbf{41.38} \\
\midrule
\midrule
\multirow{12}{*}{\raisebox{-1.5cm}{\hspace{-1em}\rotatebox[origin=c]{90}{SmoothQuant}}}
& NF4 & \textbf{72.50} & \textbf{55.22} & \textbf{66.54} & 76.66 & 74.28 & 40.70  \\
& SF4 & 71.90 & 55.09 & 66.06 & \textbf{77.04} & \textbf{75.35} & \textbf{41.04}  \\
\midrule
& INT4 & 70.35 & 54.07 & 65.43 & 75.79 & 68.90 & 39.85  \\
\midrule
& I-E2M1 & 70.39 & 53.92 & \textbf{66.22} & \textbf{76.28} & 72.11 & 39.33  \\
& B-E2M1 & 70.44 & 53.73 & 64.96 & 75.03 & 69.88 & 39.51  \\
& E2M1 & \textbf{72.21} & 55.10 & 65.9 & 76.93 & \textbf{74.71} & \textbf{41.38}  \\
& + SR & 64.25 & 47.97 & 61.33 & 73.01 & 68.96 & 34.47  \\
& + SP & 71.78 & \textbf{55.13} & 65.75 & 77.37 & 73.94 & 39.93  \\
& E3M0 & 66.74 & 51.16 & 64.25 & 75.68 & 71.71 & 36.18  \\
\midrule
& APoT4 & \textbf{71.82} & 54.87 & \textbf{66.22} & 76.39 & 73.76 & \textbf{40.36}  \\
&\texttt{+} SP & 71.78 & \textbf{55.13} & 65.75 & \textbf{77.37} & \textbf{73.94} & 39.93 \\
\bottomrule
\end{tabular}%
\vspace{-5pt}
\caption{ \textbf{LLaMA-7B W4A4 Subchannel 128}}
\label{tab:llama_qa}
\vspace{-10pt}
\end{table}
\begin{table}[t]
\centering
\footnotesize
\small
\setlength{\tabcolsep}{1.5pt}
\begin{tabular}{@{}crrrrrrrr@{}} 
\toprule
& & \textbf{LAMB} & \textbf{Hella} & \textbf{Wino} & \textbf{PIQA} & \textbf{BoolQ} & \textbf{ARC-c} \\
\midrule
\multirow{12}{*}{\raisebox{-1.5cm}{\hspace{-1em}\rotatebox[origin=c]{90}{No SmoothQuant}}}
& FP32 & 75.90 & 61.22 & 73.88 & 80.58 & 83.58 & 50.43  \\
\midrule
& NF4 & 72.02 & 59.66 & 68.11 & 79.38 & 80.64 & \textbf{47.18}  \\
& SF4 & \textbf{73.47} & \textbf{59.83} & \textbf{69.38} & \textbf{79.71} & \textbf{81.10} & 46.25  \\
\midrule
& INT4 & 64.99 & 58.11 & 67.01 & 77.69 & 76.82 & 44.37  \\
\midrule
& I-E2M1 & 66.41 & 57.23 & 68.59 & 78.35 & 74.98 & 44.62  \\
& B-E2M1 & 64.22 & 57.19 & 66.22 & 77.09 & 75.29 & 42.66  \\
& E2M1 & \textbf{72.0} & 59.56 & \textbf{69.85} & \textbf{79.05} & 79.60 & \textbf{45.14}  \\
& + SR & 65.01 & 51.32 & 66.46 & 75.35 & 76.02 & 39.33  \\
& + SP & 70.83 & \textbf{59.66} & 69.30 & 78.56 & 79.57 & 44.71  \\
& E3M0 & 70.87 & 55.48 & 66.14 & 77.86 & \textbf{80.12} & 42.15  \\
\midrule
& APoT4 & \textbf{71.2} & 59.29 & 68.43 & \textbf{79.38} & 79.33 & \textbf{45.65}  \\
&\texttt{+} SP & 70.83 & \textbf{59.66} & \textbf{69.30} & 78.56 & \textbf{79.57} & 44.71 \\
\midrule
\midrule
\multirow{12}{*}{\raisebox{-1.5cm}{\hspace{-1em}\rotatebox[origin=c]{90}{SmoothQuant}}}
& NF4 & 73.86 & 59.17 & 71.19 & 79.54 & 80.58 & 46.42  \\
& SF4 & \textbf{74.50} & \textbf{59.64} & \textbf{71.74} & \textbf{79.98} & \textbf{82.20} & \textbf{46.67}  \\
\midrule
& INT4 & 68.41 & 57.91 & 68.41 & 77.89 & 77.52 & 45.76  \\
\midrule
& I-E2M1 & 68.97 & 58.54 & 68.27 & 78.56 & 76.12 & 45.05  \\
& B-E2M1 & 68.91 & 57.86 & 68.90 & 78.45 & 75.38 & 44.20  \\
& E2M1 & 73.63 & 59.45 & \textbf{71.98} & \textbf{79.92} & \textbf{79.91} & \textbf{45.90}  \\
& + SR & 64.93 & 50.29 & 65.75 & 75.3 & 72.05 & 35.58  \\
& + SP & \textbf{73.67} & \textbf{59.63} & 69.14 & 79.43 & 79.88 & 45.65  \\
& E3M0 & 71.53 & 55.82 & 66.77 & 77.09 & 79.42 & 43.09  \\
\midrule
& APoT4 & \textbf{73.67} & 59.37 & \textbf{69.69} & 78.67 & 79.42 & 46.25  \\
&\texttt{+} SP & \textbf{73.67} & \textbf{59.63} & 69.14 & \textbf{79.43} & \textbf{79.88} & \textbf{45.65} \\
\bottomrule
\end{tabular}%
\vspace{-5pt}
\caption{ \textbf{Mistral-7B W4A4 Subchannel 128}}
\label{tab:mistral_qa}
\vspace{-10pt}
\end{table}
\begin{table}[t]
\centering
\footnotesize
\small
\setlength{\tabcolsep}{1.5pt}
\begin{tabular}{@{}crrrrrrrr@{}} 
\toprule
& & \textbf{LAMB} & \textbf{Hella} & \textbf{Wino} & \textbf{PIQA} & \textbf{BoolQ} & \textbf{ARC-c} \\
\midrule
\multirow{12}{*}{\raisebox{-1.5cm}{\hspace{-1em}\rotatebox[origin=c]{90}{No SmoothQuant}}}
& FP32 & 57.89 & 41.54 & 59.51 & 71.71 & 57.83 & 23.38  \\
\midrule
& NF4 & 40.13 & 36.57 & \textbf{57.14} & 66.16 & \textbf{52.08} & \textbf{22.95}  \\
& SF4 & \textbf{41.98} & \textbf{37.27} & 55.33 & \textbf{66.54} & 51.38 & 22.78  \\
\midrule
& INT4 & 28.06 & 32.65 & 53.43 & 61.92 & 47.83 & 20.99  \\
\midrule
& I-E2M1 & 39.10 & 35.50 & 52.80 & 65.02 & 46.27 & 21.42  \\
& B-E2M1 & 36.25 & 34.28 & 54.78 & 63.33 & 45.90 & \textbf{23.29}  \\
& E2M1 & 39.82 & 36.71 & \textbf{57.14} & 65.56 & 53.06 & 22.70  \\
& + SR & 40.62 & 37.16 & 54.62 & \textbf{68.01} & 51.90 & 22.78  \\
& + SP & 37.55 & 35.66 & 56.04 & 65.89 & \textbf{54.37} & 22.70  \\
& E3M0 & \textbf{44.13} & \textbf{37.82} & 54.46 & 67.74 & 50.98 & 22.01  \\
\midrule
& APoT4 & \textbf{37.69} & 35.61 & \textbf{57.54} & 64.91 & 54.16 & 21.42  \\
&\texttt{+} SP & 37.55 & \textbf{35.66} & 56.04 & \textbf{65.89} & \textbf{54.37} & \textbf{22.70} \\
\midrule
\midrule
\multirow{12}{*}{\raisebox{-1.5cm}{\hspace{-1em}\rotatebox[origin=c]{90}{SmoothQuant}}}
& NF4 & \textbf{44.75} & \textbf{38.11} & 54.46 & \textbf{67.85} & \textbf{49.63} & \textbf{23.63}  \\
& SF4 & 43.61 & 38.02 & \textbf{57.30} & 67.41 & 49.33 & 22.78  \\
\midrule
& INT4 & 42.42 & 37.22 & 54.46 & 66.81 & 52.57 & 22.44  \\
\midrule
& I-E2M1 & 43.47 & 37.03 & 55.72 & 66.05 & 50.55 & 22.35  \\
& B-E2M1 & 43.37 & 36.99 & 56.67 & 65.94 & 50.43 & \textbf{23.63}  \\
& E2M1 & \textbf{43.64} & 37.84 & \textbf{57.85} & 67.03 & 47.55 & 22.53  \\
& + SR & 40.02 & 37.27 & 57.06 & \textbf{68.12} & \textbf{53.46} & 22.18  \\
& + SP & 40.91 & 37.77 & 57.70 & 67.85 & 51.68 & 23.12  \\
& E3M0 & 42.34 & \textbf{37.87} & 55.17 & 67.52 & 52.57 & 21.84  \\
\midrule
& APoT4 & \textbf{41.72} & \textbf{37.97} & 57.54 & \textbf{68.34} & 51.53 & \textbf{23.21}  \\
&\texttt{+} SP & 40.91 & 37.77 & \textbf{57.70} & 67.85 & \textbf{51.68} & 23.12 \\
\bottomrule
\end{tabular}%
\vspace{-5pt}
\caption{ \textbf{OPT-1B W4A4 Subchannel 128}}
\label{tab:opt1b_qa}
\vspace{-10pt}
\end{table}
\begin{table}[t]
\centering
\footnotesize
\small
\setlength{\tabcolsep}{1.5pt}
\begin{tabular}{@{}crrrrrrrr@{}} 
\toprule
& & \textbf{LAMB} & \textbf{Hella} & \textbf{Wino} & \textbf{PIQA} & \textbf{BoolQ} & \textbf{ARC-c} \\
\midrule
\multirow{12}{*}{\raisebox{-1.5cm}{\hspace{-1em}\rotatebox[origin=c]{90}{No SmoothQuant}}}
& FP32 & 67.69 & 50.49 & 65.43 & 76.28 & 66.06 & 30.72  \\
\midrule
& NF4 & 64.89 & \textbf{47.86} & 62.75 & \textbf{74.54} & \textbf{63.21} & \textbf{29.01}  \\
& SF4 & \textbf{65.57} & 47.81 & \textbf{63.54} & 74.37 & 62.20 & 27.99  \\
\midrule
& INT4 & 53.15 & 44.98 & 60.46 & 72.8 & 62.84 & 28.50  \\
\midrule
& I-E2M1 & 62.41 & \textbf{47.76} & 60.69 & 73.99 & 62.60 & 29.18  \\
& B-E2M1 & 60.39 & 47.04 & 61.01 & 73.78 & 63.00 & 29.18  \\
& E2M1 & \textbf{65.22} & 47.39 & \textbf{62.75} & \textbf{74.32} & \textbf{64.10} & 29.01  \\
& + SR & 62.47 & 46.09 & 59.67 & 73.99 & 63.52 & 27.82  \\
& + SP & 61.73 & 47.28 & 62.04 & 73.88 & 63.82 & \textbf{30.03}  \\
& E3M0 & 57.23 & 45.32 & 60.77 & 72.74 & 62.94 & 28.58  \\
\midrule
& APoT4 & 61.40 & \textbf{47.56} & \textbf{62.43} & \textbf{75.14} & 63.39 & 29.95  \\
&\texttt{+} SP & \textbf{61.73} & 47.28 & 62.04 & 73.88 & \textbf{63.82} & \textbf{30.03} \\
\midrule
\midrule
\multirow{12}{*}{\raisebox{-1.5cm}{\hspace{-1em}\rotatebox[origin=c]{90}{SmoothQuant}}}
& NF4 & 67.79 & 49.22 & 63.06 & \textbf{75.24} & \textbf{65.38} & 30.03  \\
& SF4 & \textbf{68.29} & \textbf{49.24} & \textbf{63.85} & 75.14 & 64.74 & \textbf{30.46}  \\
\midrule
& INT4 & 66.72 & 48.8 & 63.22 & 74.10 & 62.57 & 29.10  \\
\midrule
& I-E2M1 & 65.55 & 48.64 & 62.83 & 74.59 & \textbf{65.29} & 30.03  \\
& B-E2M1 & 65.94 & 48.40 & 61.72 & 74.27 & 63.06 & 30.12  \\
& E2M1 & \textbf{68.27} & 49.23 & 63.69 & \textbf{75.19} & 64.71 & \textbf{30.63}  \\
& + SR & 64.62 & 46.36 & 60.22 & 74.81 & 64.37 & 28.41  \\
& + SP & 67.75 & \textbf{49.64} & \textbf{64.25} & 74.81 & 62.87 & 30.08  \\
& E3M0 & 61.96 & 47.30 & 60.93 & 73.50 & 62.6 & 28.33  \\
\midrule
& APoT & 67.26 & 49.56 & 64.09 & \textbf{75.30} & 62.32 & \textbf{30.38}  \\
&+ SP & \textbf{67.75} & \textbf{49.64} & \textbf{64.25} & 74.81 & \textbf{62.87} & 30.08 \\
\bottomrule
\end{tabular}%
\vspace{-5pt}
\caption{ \textbf{OPT-6B W4A4 Subchannel 128}}
\label{tab:opt6b_qa}
\vspace{-10pt}
\end{table}
\begin{table}[t!]
\centering
\footnotesize
\small
\setlength{\tabcolsep}{1.5pt}
\begin{tabular}{@{}crrrrrrrr@{}} 
\toprule
& & \textbf{LAMB} & \textbf{Hella} & \textbf{Wino} & \textbf{PIQA} & \textbf{BoolQ} & \textbf{ARC-c} \\
\midrule
\multirow{12}{*}{\raisebox{-1.5cm}{\hspace{-1em}\rotatebox[origin=c]{90}{No SmoothQuant}}}
& FP32 & 62.57 & 55.84 & 75.45 & 78.78 & 83.21 & 52.56  \\
\midrule
& NF4 & 52.20 & \textbf{51.63} & 71.03 & \textbf{76.93} & 74.62 & 49.74  \\
& SF4 & \textbf{53.06} & 51.22 & \textbf{71.82} & 75.08 & \textbf{79.88} & \textbf{50.60}  \\
\midrule
& INT4 & 41.18 & 47.4 & 67.48 & 74.37 & 66.97 & 46.16  \\
\midrule
& I-E2M1 & 43.18 & 47.4 & 67.01 & 75.35 & 66.73 & 45.99  \\
& B-E2M1 & 39.82 & 46.5 & 67.88 & 74.43 & 66.64 & 42.92  \\
& E2M1 & 49.66 & \textbf{51.19} & 71.82 & 75.30 & 78.29 & 49.23  \\
& +SR & \textbf{51.81} & 49.40 & \textbf{73.56} & 75.73 & \textbf{78.47} & 47.10  \\
& + SP & 51.19 & 50.85 & 69.46 & \textbf{76.50} & 77.58 & \textbf{49.32}  \\
& E3M0 & 42.15 & 47.63 & 66.61 & 74.05 & 72.81 & 45.22  \\
\midrule
& APoT4 & 49.58 & 50.25 & \textbf{69.85} & \textbf{76.77} & 75.60 & 48.46  \\
&\texttt{+} SP & \textbf{51.19} & \textbf{50.85} & 69.46 & 76.50 & \textbf{77.58} & \textbf{49.32} \\
\midrule
\midrule
\multirow{12}{*}{\raisebox{-1.5cm}{\hspace{-1em}\rotatebox[origin=c]{90}{SmoothQuant}}}
& NF4 & 52.98 & \textbf{51.74} & \textbf{71.82} & 75.73 & 79.72 & 49.23  \\
& SF4 & \textbf{55.33} & 51.53 & \textbf{71.82} & \textbf{76.44} & \textbf{80.92} & \textbf{49.74}  \\
\midrule
& INT4 & 31.94 & 46.57 & 64.96 & 72.03 & 69.45 & 44.54  \\
\midrule
& I-E2M1 & 36.97 & 47.85 & 67.88 & 72.63 & 67.37 & 46.50  \\
& B-E2M1 & 31.13 & 45.91 & 64.56 & 72.58 & 66.97 & 40.70  \\
& E2M1 & 51.68 & \textbf{51.33} & 71.03 & 76.28 & 77.92 & \textbf{50.17}  \\
& + SR & \textbf{52.78} & 49.39 & 72.93 & \textbf{76.39} & 78.01 & 48.21  \\
& + SP & 49.95 & 50.86 & \textbf{71.74} & 74.92 & \textbf{81.25} & 48.38  \\
& E3M0 & 49.41 & 47.51 & 69.14 & 74.70 & 71.41 & 44.88  \\
\midrule
& APoT4 & 47.86 & 50.49 & 70.40 & \textbf{75.14} & 79.11 & 47.53  \\
&\texttt{+} SP & \textbf{49.95} & \textbf{50.86} & \textbf{71.74} & 74.92 & \textbf{81.25} & \textbf{48.38} \\
\bottomrule
\end{tabular}%
\vspace{-5pt}
\caption{ \textbf{Phi-2 W4A4 Subchannel 128}}
\label{tab:phi_qa}
\vspace{-10pt}
\end{table}

\end{document}